\documentclass{article} % For LaTex2e
\usepackage{iclr2025_conference,times}

\usepackage{amsmath,amsfonts,bm}

\def\eqref#1{equation~\ref{#1}}
\def\1{\bm{1}}

\DeclareMathAlphabet{\mathsfit}{\encodingdefault}{\sfdefault}{m}{sl}
\SetMathAlphabet{\mathsfit}{bold}{\encodingdefault}{\sfdefault}{bx}{n}

\usepackage{hyperref}
\usepackage{url}

\usepackage{enumitem}
\usepackage{amsmath}
\usepackage{multirow}
\usepackage{adjustbox}
\usepackage{array}
\usepackage{booktabs} % for professional tables
\usepackage{tabularx}
\usepackage{graphicx}

\usepackage{amssymb}
\usepackage{pifont}
\newcommand{\cmark}{\ding{51}}%
\newcommand{\xmark}{\ding{55}}%

\newcommand\blfootnote[1]{%
\begingroup
\renewcommand\thefootnote{}\footnote{#1}%
\addtocounter{footnote}{-1}%
\endgroup
}

\usepackage[most]{tcolorbox}

\definecolor{codegreen}{rgb}{0,0.6,0}
\definecolor{codegray}{rgb}{0.5,0.5,0.5}
\definecolor{codepurple}{rgb}{0.58,0,0.82}
\definecolor{backcolour}{rgb}{0.95,0.95,0.92}
\usepackage{listings}
\lstdefinestyle{mystyle}{
  commentstyle=\color{codegreen},
  keywordstyle=\color{magenta},
  numberstyle=\small\color{codegray},
  stringstyle=\color{codepurple},
  basicstyle=\small,
  breakatwhitespace=false,         
  breaklines=true,                 
  captionpos=b,                    
  keepspaces=false,                                 
  showspaces=false,                
  showstringspaces=false,
  showtabs=false,                  
  tabsize=2
}

\title{MathCoder2: Better Math Reasoning from Continued Pretraining on Model-translated Mathematical Code}

\author{Zimu Lu$^{*}$, Aojun Zhou$^{*}$, Houxing Ren, Ke Wang, Weikang Shi\\ {\bf Junting Pan}, {\bf Mingjie Zhan}$^{\dagger}$, {\bf Hongsheng Li}$^{\dagger}$\\
  Multimedia Laboratory (MMLab), The Chinese University of Hong Kong\\
 \texttt{luzimu@mail.ustc.edu.cn} \quad \texttt{\{aojunzhou, zmjdll\}@gmail.com}  \\ \texttt{hsli@ee.cuhk.edu.hk} 
}

\begin{document}

\maketitle

\begin{abstract}
Code has been shown to be effective in enhancing the mathematical reasoning abilities of large language models due to its precision and accuracy. Previous works involving continued mathematical pretraining  often include code that utilizes math-related packages, which are primarily designed for fields such as engineering, machine learning, signal processing, or module testing, rather than being directly focused on mathematical reasoning. In this paper, we introduce a novel method for generating mathematical code accompanied with corresponding reasoning steps for continued pretraining. Our approach begins with the construction of a high-quality mathematical continued pretraining dataset by incorporating math-related web data, code using mathematical packages, math textbooks, and synthetic data. Next, we construct reasoning steps by extracting LaTeX expressions, the conditions needed for the expressions, and the results of the expressions from the previously collected dataset. Based on this extracted information, we generate corresponding code to accurately capture the mathematical reasoning process. Appending the generated code to each reasoning step results in data consisting of paired natural language reasoning steps and their corresponding code. Combining this data with the original dataset results in a 19.2B-token high-performing mathematical pretraining corpus, which we name MathCode-Pile. Training several popular base models with this corpus significantly improves their mathematical abilities, leading to the creation of the MathCoder2 family of models. All of our data processing and training code is open-sourced, ensuring full transparency and easy reproducibility of the entire data collection and training pipeline. The code is released at~\url{https://github.com/mathllm/MathCoder2}.
\end{abstract}

\blfootnote{$^*$Equal contribution\quad $^\dagger$Corresponding author}

\section{Introduction}

Various studies~\citep{azerbayev2024llemmaopenlanguagemodel,shao2024deepseekmathpushinglimitsmathematical} have shown that training on code enhances the mathematical reasoning abilities of large language models (LLMs). Previous research in continued mathematical pretraining often includes code that utilizes math-related packages~\citep{azerbayev2024llemmaopenlanguagemodel}. This code is typically sourced from GitHub and is primarily designed for fields such as engineering, machine learning, signal processing, or module testing, rather than focusing directly on mathematics. Recent models~\citep{zhou2024solving,yang2024qwen25mathtechnicalreportmathematical,ying2024internlmmath,shao2024deepseekmathpushinglimitsmathematical,wang2023mathcoderseamlesscodeintegration} have adopted Tool-Integrated Reasoning (TIR) in fine-tuning. They utilize integrated natural language reasoning steps and Python code to improve performance on mathematical reasoning tasks. Reasoning with the help of code is particularly effective for more challenging problems, likely due to its precision and accuracy. 

Although utilizing existing open-source code in the pretraining phase can enhance the mathematical reasoning abilities of LLMs, such code often lacks accompanying natural language explanations or context. This might hinder the model's ability to fully understand them. In this paper, we propose a novel method for \textit{generating large amounts  of mathematical code accompanied by corresponding natural language reasoning steps}, which are extracted from math-related pretraining texts. Different from the existing math-related code, our generated code is paired with natural language reasoning steps, making the code more comprehensible. Also, as our code is generated based on math-related texts, they are all highly related to mathematical reasoning. When used in pretraining, the mathematical code paired with reasoning steps facilitates LLMs' understanding of math-related pretraining texts, as it effectively captures the underlying reasoning process. Furthermore, this data enhances the model's potential to be finetuned for TIR reasoning.

Our data processing pipeline consists of two key steps: (1) carefully curating a robust basic dataset for pretraining, and (2) generating paired reasoning steps and mathematical code by extracting LaTeX expressions and their context, translating the extracted information into Python code snippets, executing the generated code snippets, and verifying their correctness.

First, we gather and carefully filter a wide variety of math-related data sources, including web pages, model-generated data, math-related code, and textbooks. Through an advanced filtering process, we ensure the dataset is both large and highly relevant, minimizing irrelevant content while preserving the mathematical texts necessary for training. This results in a 16.5B-token dataset that forms the foundation of our pretraining efforts. By conducting experiments with smaller models, we show that this careful curation leads to more efficient training without sacrificing model performance.

\begin{figure*}[t]
    \centering
    \includegraphics[width=1.0\textwidth]{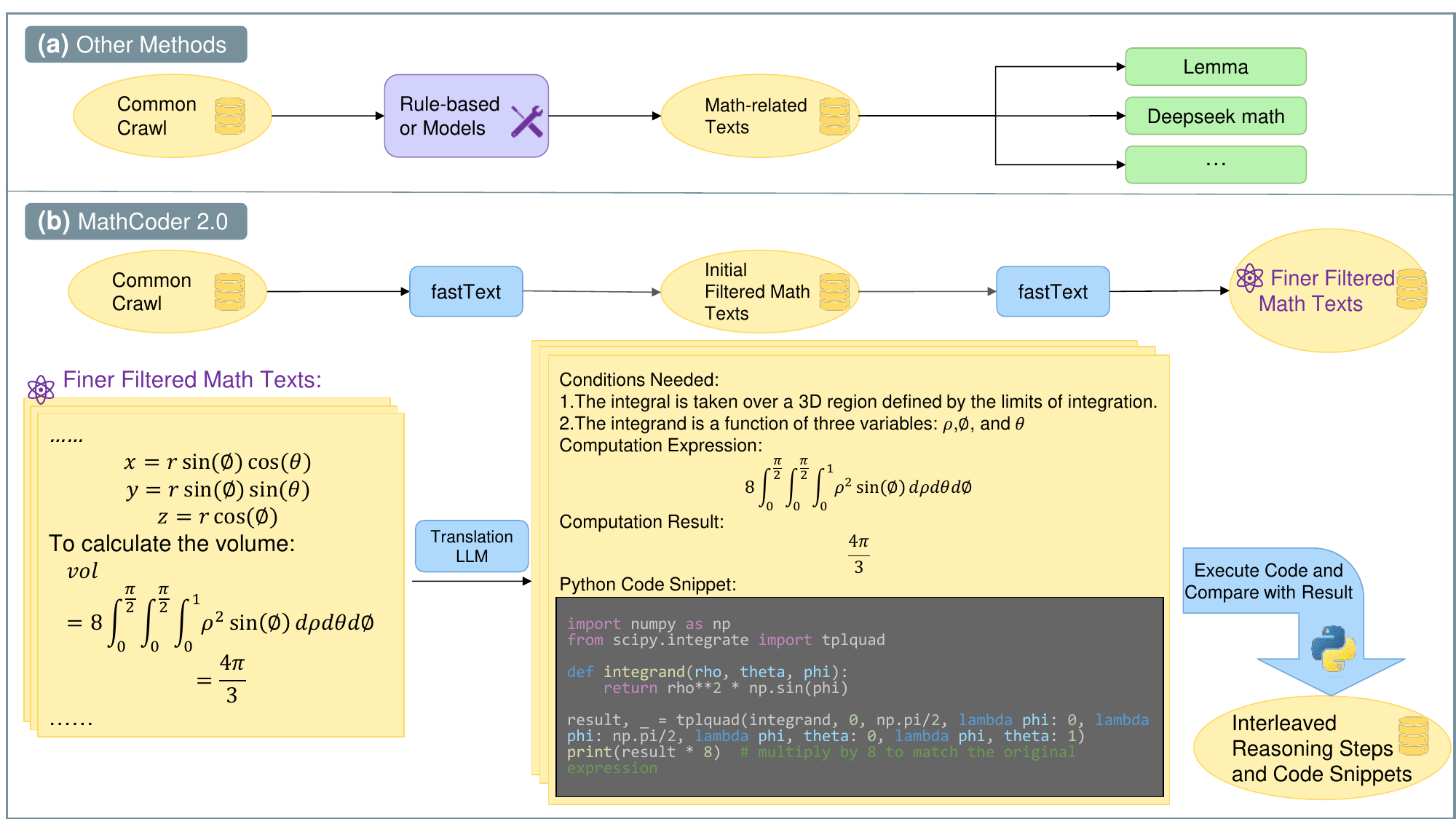}
    \caption{The data processing pipeline. (a) shows the pipeline of prior works. (b) demonstrates our method. We first use a fastText classifier to filter the Common Crawl corpus, resulting in the initial filtered math texts. Then, we  annotate part of the filtered texts to train a new fastText classifier, and conduct a second filtering, resulting in the finer filtered math texts. Then we use an instruction-tuned LLM to extract reasoning steps from these math-related texts, and translate the reasoning steps into corresponding code snippets. We execute the code snippets and compare the output with the expected result. If the code executes successfully and the result is as expected, the code is retained.}
    
\label{fig:pipeline}
\end{figure*}

Second, we propose a novel method for generating large amounts of paired mathematical reasoning steps and their corresponding Python code. Given a piece of text from the pretraining corpus collected above, we wrap it in a carefully designed prompt that instructs a Llama-3.1-70B-Instruct model to extract LaTeX expressions along with their relevant context, including the conditions for each expression and the result of its computation. This results in a list of comprehensive mathematical reasoning steps, complete with the necessary conditions, the computations taken, and the results. Then, we prompt the model to translate each reasoning step into a Python code snippet that captures the underlying reasoning process. The generated Python snippets are executed, and only those that run successfully and produce outputs matching the expected results are retained. By pairing the code with the corresponding reasoning step, we create the final data. The process is demonstrated in the lower half of Fig.~\ref{fig:pipeline}. This process yields a 2.7B-token corpus of mathematical code snippets accompanied with their corresponding reasoning steps, which we combine with the data generated in the first step to create a 19.2B-token pretraining dataset, named \textbf{MathCode-Pile}.

We validate the effectiveness of MathCode-Pile on four popular base models: Llama-3-8B, DeepSeekMath-7B, Mistral-7B, and Code-Llama-7B, significantly improving their performance on five representative mathematical benchmarks. We name the resulting family of pretrained models MathCoder2. In particular, MathCoder2-Llama-3-8B achieves 4-shot accuracies of 38.4\% on MATH and 69.9\% on GSM8K, outperforming the baseline of training only on the basic data generated in the first step by 3.1\% and 4.1\%, respectively. This demonstrates that the data of mathematical code accompanied with reasoning steps effectively enhances LLMs' reasoning abilities.

Different from recent works, such as DeepSeekMath~\citep{shao2024deepseekmathpushinglimitsmathematical}, InternLM-Math~\citep{ying2024internlmmath}, and Qwen2.5-Math~\citep{yang2024qwen25mathtechnicalreportmathematical}, which only release their model weights, we offer a detailed, open-source framework for data processing and training that achieves performance competitive with these models, fostering further progress in mathematical reasoning for LLMs.

Our contributions include:

\begin{itemize}
    \item A novel and effective method for generating large amounts of mathematical code with corresponding natural language reasoning steps, significantly enhancing pretraining outcomes.
    \item The creation of MathCode-Pile, a meticulously curated 19.2B-token dataset for continued mathematical pretraining. This dataset includes math-related web data, synthetic data, code, textbooks, and model-translated mathematical code.
    \item Full open-sourcing of all data processing and training code, ensuring transparency and reproducibility to support future research.
\end{itemize}

\section{Curation of MathCode-Pile}

We curate our mathematical pretraining dataset, MathCode-Pile, in two steps: first, we collect the basic data in Sec.~\ref{sec:basic_data}, and then we use it to generate mathematical code snippets with their corresponding natural language reasoning steps in Sec.~\ref{sec:mathematical_code}.

\subsection{Basic Data}
\label{sec:basic_data}
We collect and carefully filter a diverse range of mathematical data to ensure relevance and quality for continued pretraining of LLMs. The data includes math-related web content, synthetic data, code utilizing mathematical packages, and mathematical textbooks.

\textbf{Math-related Web Data}. Web data offers a broad range of real-world mathematical examples. We start with the OpenWebMath~\citep{paster2023openwebmathopendatasethighquality} dataset, which contains mathematical web pages sourced from Common Crawl. Observing that a significant portion of these documents are unrelated to mathematics, we instruct the Mixtral-8x7B-Instruct model with a carefully designed prompt (detailed in Appendix~\ref{sec:classify_prompt}) to filter out irrelevant texts. Examples of irrelevant texts are shown in Appendix~\ref{sec:irrelevant_texts}. This reduces the dataset from 13.7B tokens to 4.8B tokens (measured using the Llama-3 tokenizer). We call this filtered version filtered-OpenWebMath.

To further expand the dataset, we train a fastText classifier~\citep{joulin2016bagtricksefficienttext} using filtered-OpenWebMath as positive samples and random Common Crawl data as negative samples (training details are explained in Appendix.~\ref{sec:fasttext}). This model helps identify additional math-related documents within the Common Crawl data from Matrix~\citep{zhang2024mapneohighlycapabletransparent}, a general pretraining dataset. A second round of filtering is performed, where Mixtral-8x7B-Instruct annotates a portion of these documents, and a new fastText classifier trained based on these annotations further refines the data. This produces 6.4B tokens, which we label as filtered-CC-En-math. Finally, we combine filtered-OpenWebMath and filtered-CC-En-math, resulting in a comprehensive 11.2B-token math-related web dataset.

\textbf{Synthetic Data}. Synthetic data offers structured mathematical texts that complement the web data. We collect synthetic data from various open-source repositories on Hugging Face, including datasets like Education-College-Students\footnote{https://huggingface.co/datasets/ajibawa-2023/Education-College-Students}, Maths-College\footnote{https://huggingface.co/datasets/ajibawa-2023/Maths-College}, and synthetic math books from Matrix~\citep{zhang2024mapneohighlycapabletransparent}. To ensure relevance, we apply a fastText classifier to filter out non-mathematical documents, refining the dataset to 2.2B tokens of high-quality synthetic math content.

\textbf{Code Utilizing Mathematical Packages}. Code data offers practical examples of how mathematical libraries are used in programming. We collect code from Python and Jupyter files within the StarCoderData dataset~\citep{li2023starcodersourceyou}, retaining only programs that import math-related packages such as sympy, fractions, cmath, scipy, or statistics. The widely used numpy package is not used to filter the documents, as it appears frequently in non-mathematical contexts. After filtering, this collection process results in 1.7B tokens of code related to mathematical computations.

\textbf{Mathematical Textbooks}. Textbooks provide formal, structured presentations of mathematical concepts, making them a valuable source of math knowledge. We gather 8K PDFs of textbooks from online resources by identifying those with titles containing math-related keywords such as algebra, geometry, probability, etc. These PDF files are then converted into markdown format using the Nougat tool for easier integration into our training pipeline.

\subsection{Model-translated Mathematical Code}
\label{sec:mathematical_code}
\begin{figure}
\begin{tcolorbox}[colback=blue!5!white,colframe=blue!75!black]
\begin{small}
\textbf{Prompt:}
You will be presented with a text related to math. I need you to identify all the complex computations in it. For each complex computation, find out the conditions needed for the computation, the LaTeX expression that conducts the computation, and the result of the computation. Then generate a Python code snippet for each computation that demonstrates how the result is reached. Output each computation in the following format:

\vspace{2mm}

Conditions Needed:
\vspace{0.05mm}

1. [Condition 1]
\vspace{0.05mm}

2. [Condition 2]
\vspace{0.05mm}

...

\vspace{2mm}

Computation Expression:
\vspace{0.05mm}

\$[LaTeX Expression]\$

\vspace{2mm}

Computation Result:
\vspace{0.05mm}

[Computation Result]

\vspace{2mm}

Python Code Snippet:
\vspace{0.05mm}
\begin{verbatim}
```python\end{verbatim}
[Python Code]
\begin{verbatim}```
\end{verbatim}

There can be more than one complex computation in the text. Output only the computations that requires calculation. Do not include mathematical statements or definitions as a computation. Make sure each snippet can be executed individually. The text is as follows: \{TEXT\}

The computations are:

\end{small}
\end{tcolorbox}
\caption{The prompt for extracting reasoning steps from texts in the pretraining corpus and generating the corresponding Python snippets. \{TEXT\} is replaced with the text from the dataset collected in Sec.~\ref{sec:basic_data}.}\label{fig:prompt_translation}
\end{figure}

In this section, we propose a novel approach for extracting reasoning steps from the basic pretraining data and translating them into corresponding Python code snippets that capture the underlying reasoning processes. This extraction and translation process is performed using a strong instruction-tuned model, which is Llama-3.1-70B-Instruct in this paper.

Our method begins with taking a piece of text from the basic pretraining data and wrapping it in a carefully designed prompt, as shown in Fig.~\ref{fig:prompt_translation}. This prompt instructs the model to identify \textit{LaTeX expressions} denoting complex computations, along with the necessary context, including the \textit{conditions required for the computation} and the \textit{expected result}. By explicitly extracting the conditions of the LaTeX expression, we enhance the model’s ability to comprehend the underlying mathematical reasoning behind the usage of the expression. The expected result of the computation can later serve as a basis for verifying the correctness of the generated code. A mathematical reasoning step is constructed by combining the conditions, expression and result. The prompt then directs the model to produce a \textit{Python code snippet} that accurately reflects the underlying reasoning process behind the extracted reasoning step. The model is asked to present the conditions, LaTeX expression, result, and Python code snippet in a structured format, ensuring that each part can be easily extracted from the generated text. Examples of generated texts are shown in Appendix~\ref{sec:examples_translated_code}.

After the Python code snippets are generated, they are executed, and outputs of the execution are compared with the expected results extracted from the generated text. Only the Python code snippets that execute without errors and produce correct outputs are retained. This filtering process ensures a higher quality of generated code, making the resulting dataset more reliable for mathematical pretraining compared to approaches that rely on unverified and general-purpose code.

Leveraging the Llama-3.1-70B-Instruct model, we initially generated 3.1B tokens of the data. After applying the filtering process, we obtain a total of 2.7B tokens of high-quality data of mathematical code snippets accompanied with their corresponding reasoning steps. This newly generated data significantly enriches our original pretraining corpus. By combining this data with the basic pretraining data, we create a comprehensive pretraining dataset totaling 19.2B tokens, which we refer to as \textbf{MathCode-Pile}. Detailed statistics of MathCode-Pile are presented in Tab.~\ref{tab:dataset_statistics}. This dataset is tailored specifically for enhancing the mathematical and coding abilities of LLMs. To avoid benchmark contamination, we follow~\cite{yang2024qwen25mathtechnicalreportmathematical} to filter out samples that have significant overlaps with any of the questions from benchmark datasets used in evaluation. We use exact match to remove the identical samples and further apply 13-gram deduplication (with a condition that the Jaccard similarity should be larger than 0.6) to remove more samples that might cause contamination.

\begin{table}[t]\fontsize{9}{11}\selectfont
\centering
\caption{The components and data statistics of MathCode-Pile.}
\begin{tabularx}{\columnwidth}{>{\raggedright\arraybackslash\hsize=1.3\hsize}X |>{\centering\arraybackslash\hsize=.5\hsize}X >{\centering\arraybackslash\hsize=.5\hsize}X >{\centering\arraybackslash\hsize=.5\hsize}X >{\centering\arraybackslash\hsize=.8\hsize}X}
\toprule
\textbf{Components} & \textbf{Size (MB)} & \textbf{Documents} & \textbf{Tokens} &  \textbf{Average (Tokens)} \\
\midrule
Filtered-OpenWebMath &  16,999 & 2,824,705 & 4,826,902,621 & 1,709  \\
Filtered-CC-En-math & 23,465 & 7,597,718 & 6,341,745,645 & 835  \\
Synthetic data & 8,855 & 2,195,974 & 2,193,189,314 & 999 \\
Code using math packages &  6,120 & 513,059 & 1,703,226,005 & 3,320  \\
Mathematical textbooks & 4,431 & 8,373 & 1,390,268,773 & 166,042  \\
\textbf{Translated mathematical code} & 8,235 & 6,347,823 & 2,728,740,985 & 430  \\
\midrule
Total & 68,105 & 19,487,652 & 19,184,073,343 & 984  \\
\bottomrule
\end{tabularx}

\label{tab:dataset_statistics}
\end{table}

In comparison to traditional methods of curating math-related code, which often draw on general-purpose repositories, our method ensures that the code is not only syntactically correct but also mathematically sound, reflecting a deeper understanding of mathematical reasoning. Our mathematical code is accompanied with corresponding natural language reasoning steps, which makes understanding the reasoning process easier. This makes MathCode-Pile a superior resource for models aimed at performing advanced mathematical reasoning tasks.

\section{Experiments}

To demonstrate the effectiveness of our method, we first train several base models ranging from 7B to 8B parameters using MathCode-Pile and compare them to other best-performing models of the same size. The group of models resulting from the continued mathematical pretraining is named \textbf{MathCoder2}. Next, we train and compare various other open-source math pretraining datasets against MathCode-Pile using a smaller model, DeepSeekCoder-1.3B. To showcase the potential of the MathCoder2 models, we further perform supervised fine-tuning on them. Finally, we conduct ablation studies to analyze the impact of each component of the dataset.

\subsection{Main Results}

\begin{table}[t]\fontsize{9}{11}\selectfont
\centering
\caption{Performance of various pretrained models on five representative mathematical datasets. All results reported are based on greedy decoding. ``Code-open'' shows whether the code for data-processing and model-training is open-sourced. The {\color{red}red} numbers show the improvements compared to the base model from which each MathCoder2 model is trained.}
\begin{tabularx}{\columnwidth}{>{\raggedright\arraybackslash\hsize=1.9\hsize}X >{\centering\arraybackslash\hsize=.3\hsize}X >{\centering\arraybackslash\hsize=.3\hsize}X *{4}{>{\centering\arraybackslash\hsize=.5\hsize}X} >{\centering\arraybackslash\hsize=.5\hsize}X}
\toprule
\textbf{Model} & \textbf{Size} & \textbf{Code-open} & \textbf{MATH} &   \textbf{GSM8K}  & \textbf{SAT} &  \textbf{OCW} & \textbf{MMLU-MATH} \\
\midrule
Qwen2-Math & 7B & \xmark & 50.4 & 80.4 & 87.5 & 14.0 & 57.9 \\
Qwen2.5-Math & 7B & \xmark & 55.4 & 91.6 & - & - & - \\
InternLM2.5 & 7B & \xmark & 34.0 & 74.8 & 65.6 & 8.1 & 49.6 \\
InternLM2-Math-Base & 7B & \xmark & 21.5 & 49.2 & - & - & - \\
Llemma & 7B & \cmark & 18.0 & 36.4 & 53.1 & 7.7 & - \\
Llama-2 & 7B & \xmark & 3.2 & 11.8 & 25.0 &  3.7 & - \\
\midrule
Llama-3 & 8B & \xmark & 21.4 & 54.8 & 56.3 & 10.3 & 42.8 \\
\textbf{MathCoder2-Llama-3} & 8B & \cmark & 38.4\tiny\color{red}({+17.0}) & 69.9\tiny\color{red}({+15.1}) & 84.4\tiny\color{red}({+28.1}) & 18.0\tiny\color{red}({+7.7}) & 46.5\tiny\color{red}({+3.7}) \\
\midrule
DeepSeekMath & 7B & \xmark & 36.2 & 64.2 & 84.4 & 15.4 & 47.4 \\
\textbf{MathCoder2-DeepSeekMath} & 7B & \cmark & 38.6\tiny\color{red}({+2.4}) & 68.8\tiny\color{red}({+4.6}) & 90.6\tiny\color{red}({+6.2}) & 16.9\tiny\color{red}({+1.5}) & 48.3\tiny\color{red}({+0.9}) \\
\midrule
Mistral & 7B & \xmark & 13.1 & 52.2 & 75.0 & 8.5 & 38.3 \\
\textbf{MathCoder2-Mistral} & 7B & \cmark & 36.7\tiny\color{red}({+23.6}) & 68.2\tiny\color{red}({+16.0}) & 81.3\tiny\color{red}({+6.3}) & 13.2\tiny\color{red}({+4.7}) & 42.2\tiny\color{red}({+3.9}) \\
\midrule
Code-Llama & 7B & \xmark & 6.7 & 14.6 & 25.0 & 3.7 & 26.4 \\
\textbf{MathCoder2-Code-Llama} & 7B & \cmark & 28.8\tiny\color{red}({+22.1}) & 52.3\tiny\color{red}({+37.7}) & 71.9\tiny\color{red}({+46.9}) & 8.5\tiny\color{red}({+4.8}) & 33.7\tiny\color{red}({+7.3}) \\
\bottomrule
\end{tabularx}

\label{tab:main_results}
\end{table}

\textbf{Benchmark datasets}. We evaluate the MathCoder2 models on five representative datasets: GSM8K~\citep{cobbe2021trainingverifierssolvemath}, MATH~\citep{hendrycks2021measuringmathematicalproblemsolving}, SAT-Math~\citep{azerbayev2024llemmaopenlanguagemodel}, OCW~\citep{lewkowycz2022solvingquantitativereasoningproblems}, and MMLU-Math~\citep{hendrycks2021measuringmassivemultitasklanguage}. GSM8K and MATH are tested using a 4-shot prompt with MAmmoTH's evaluation framework~
\citep{yue2023mammothbuildingmathgeneralist}. SAT-Math and OCW are tested using a 4-shot prompt with DeepSeekMath's evaluation framework~\citep{shao2024deepseekmathpushinglimitsmathematical}. MMLU-Math is tested using the lm-evaluation-harness's~\citep{eval-harness} default zero-shot setting for MMLU. These datasets cover a wide range of mathematical problems across various types and difficulty levels, from primary school math word problems to college-level challenges, providing a comprehensive evaluation of the models.

\textbf{Base models and training settings}. To demonstrate that our pretraining corpus is effective across different base models, we continue pretraining four base models with MathCode-Pile: Llama-3-8B~\citep{dubey2024llama}, DeepSeekMath-7B~\citep{shao2024deepseekmathpushinglimitsmathematical}, Mistral-7B~\citep{jiang2023mistral7b}, and Code-Llama-7B~\citep{roziere2024codellamaopenfoundation}. MathCoder2-Llama-3-8B is trained for 3 epochs with a global batch size of 4 million tokens and an 8192 token context length. MathCoder2-DeepSeekMath, MathCoder2-Mistral, and MathCoder2-CodeLlama are each trained for 3 epochs with a global batch size of 4 million tokens and a 4096 token context length.

\textbf{Baselines}. We compare our method with various other base models that possess strong mathematical abilities and are of similar sizes, including Qwen2-Math 7B~\citep{yang2024qwen2}, Qwen2.5-Math 7B~\citep{yang2024qwen25mathtechnicalreportmathematical}, InternLM2-Math-Base 7B~\citep{ying2024internlmmath}, InternLM2.5 7B~\citep{cai2024internlm2}, DeepSeekMath 7B~\citep{shao2024deepseekmathpushinglimitsmathematical}, Llemma 7B~\citep{azerbayev2024llemmaopenlanguagemodel}, Mistral 7B~\citep{jiang2023mistral7b}, Llama2 7B~\citep{touvron2023llama2openfoundation}, Llama3 8B~\citep{dubey2024llama} and Code-Llama 7B~\citep{roziere2024codellamaopenfoundation}.

\textbf{Results}: As demonstrated in Tab.~\ref{tab:main_results}, continued pretraining on MathCode-Pile consistently improves performance across all five benchmark datasets. MathCoder2 models rival the performance of top models like InternLM2-Math-Base, InternLM2.5, and DeepSeekMath. In particular, MathCoder2-DeepSeekMath demonstrates that our method continues to enhance the performance of DeepSeekMath, a model that has already been extensively trained on large amounts of math-related data. However, there remains a performance gap between MathCoder2 and the Qwen2-Math and Qwen2.5-Math models. This gap might be attributed to their superior computational, manual, and financial resources, which enable the scaling of data size and the further improvement of data quality, reporting a mathemtaical dataset of 700B tokens~\citep{yang2024qwen25mathtechnicalreportmathematical}.

In contrast to models like Qwen2-Math, which only open-source their model weights, with much of their data processing and training details undisclosed, MathCoder2 is fully open-sourced, including all data processing pipelines and training code. This openness facilitates transparency, reproducibility, and further research, which is crucial for advancing the field. Compared to Llemma, which also open-sources its code, our method achieves better performance on the five datasets. Particularly, when trained on the same base model, Code-Llama, our method performs significantly better, which demonstrates the effectiveness of the MathCode-Pile pretraining corpus.

\subsection{Post-training}

To further demonstrate the potential of the MathCoder2 models in aligning to mathematical problem-solving tasks, we select the MathCoder2-Llama-3-8B model and MathCoder2-DeepSeekMath-7B for finetuning on mathematical problem-solution pairs. We first train the base model on general mathematical instructions following~\cite{yue2024mammoth2scalinginstructionsweb} for two epochs. Subsequently, we finetune the model on NuminaMath-CoT\footnote{https://huggingface.co/datasets/AI-MO/NuminaMath-CoT}, and NuminaMath-TIR\footnote{https://huggingface.co/datasets/AI-MO/NuminaMath-TIR} datasets for three epochs.

\begin{table}[t]\fontsize{9}{11}\selectfont
\centering
\caption{Performance of various finetuned models on five representative mathematical datasets. All results reported are based on greedy decoding.}
\begin{tabularx}{\columnwidth}{>{\raggedright\arraybackslash\hsize=2.6\hsize}X >{\centering\arraybackslash\hsize=.2\hsize}X *{2}{>{\centering\arraybackslash\hsize=.4\hsize}X}  >{\centering\arraybackslash\hsize=.3\hsize}X >{\centering\arraybackslash\hsize=.6\hsize}X *{2}{>{\centering\arraybackslash\hsize=.4\hsize}X}}
\toprule
\textbf{Model} & \textbf{Size} & \textbf{MATH} &   \textbf{GSM8K}  & \textbf{OCW} &  \textbf{Olympiad Bench}  & \textbf{SVAMP} \\
\midrule
Qwen2-Math-Instruct & 7B & 75.1 & 89.9 & 34.6 & 38.2 &  - \\
Qwen2.5-Math-Instruct & 7B & 83.6 & 95.2 & 37.1 & 41.6 &  - \\
DeepSeekMath-Instruct-CoT & 7B & 46.8 & 82.9 & - & - &  - \\
DeepSeekMath-Instruct-TIR & 7B & 57.4 & 83.7 & - & - &  - \\
InternLM2-math-plus & 7B & 54.4 & 84.0 & 17.3 & 18.8 &  - \\
NuminaMath-7B-CoT & 7B & 55.2 & 75.4 & 19.1 & 19.9 & - \\
NuminaMath-7B-TIR & 7B & 68.1 & 84.6 & - & - & - \\
ToRA-Code & 7B & 44.6 & 72.6 & - & - & 70.4 \\
MathCoder & 7B & 30.2 & 67.8 & - & - & 70.7 \\
MAmmoTH2-Plus & 8B & 42.8 & 84.1 & - & - & - \\
Llama-3.1-Instruct & 8B & 47.2 & 76.6 & 21.7 & 15.4  & - \\
\midrule
\textbf{MathCoder2-Llama-3-Instruct-CoT} & 8B & 58.5 & 83.9 & 29.4 & 25.8 &  92.7 \\
\textbf{MathCoder2-Llama-3-Instruct-TIR} & 8B & 69.7 & 85.8 & 37.6 & 37.6 &  94.9 \\
\textbf{MathCoder2-DeepSeekMath-Instruct-CoT} & 7B & 55.2 & 80.3 & 30.9 & 23.0 &  92.1 \\
\textbf{MathCoder2-DeepSeekMath-Instruct-TIR} & 7B & 69.6 & 86.5 & 41.9 & 37.9 &  92.8 \\
\bottomrule
\end{tabularx}

\label{tab:post_training}
\end{table}

The results are shown in Tab.~\ref{tab:post_training}. MathCoder2-Instruct-TIR achieves high performance on all five datasets, reaching 69.7\% on MATH and 86.5\% on GSM8K, outperforming many of the best open-source models of similar size and demonstrating our method's potential to improve performance on downstream mathematical reasoning tasks. As this paper focuses on continued mathematical pretraining, the post-training serves only as a validation of the potential of our models. We conducted only simple supervised fine-tuning, without performing reinforcement learning or direct preference optimization, which could further improve performance on downstream tasks.

\subsection{Ablation Studies}

In this session, we first analyze the impact of various components of the training data. Next, we compare MathCode-Pile to other open-source mathematical pretraining corpora.

\begin{table}[t]\fontsize{9}{11}\selectfont
\centering
\caption{Analysis of the impact of the mathematical code. The upper half of the table presents the results of both using and not using the mathematical code data. The lower half displays an ablation study on the design of concatenating the reasoning steps and code snippets for pretraining. ``Basic + Reasoning-step-only'' represents only adding the conditions, expressions, and results, while ``Basic + Trans-code-only'' represents only adding the translated code.  ``Reasoning-Step\&Code'' represents the concatenated data combining both. ``Basic + No-code-prompt'' represents using a prompt that simply instruct Llama-3.1-70B-Instruct to rewrite texts to improve their quality.
}
\begin{tabularx}{\columnwidth}{>{\raggedright\arraybackslash\hsize=1.8\hsize}X >{\centering\arraybackslash\hsize=1.3\hsize}X *{5}{>{\centering\arraybackslash\hsize=.35\hsize}X}}
\toprule
\textbf{Data Composition} & \textbf{Base Model} & \textbf{MATH} &   \textbf{GSM8K}  & \textbf{SAT} &  \textbf{OCW} & \textbf{MMLU-MATH} \\
\midrule
Basic & Llama-3-8B & 34.7 & 65.8 & 71.9 & 12.9 & 45.2 \\
Basic + Reasoning-Step\&Code & Llama-3-8B & 38.4\tiny\color{red}({+3.7}) & 69.9\tiny\color{red}({+4.1}) & 84.4\tiny\color{red}({+12.5}) & 18.0\tiny\color{red}({+5.1}) & 46.5\tiny\color{red}({+1.3}) \\
\midrule
Basic + Reasoning-step-only & DeepSeekCoder-1.3B & 16.7 & 22.7 & 40.6 & 4.8 & 25.9 \\
Basic + Trans-code-only & DeepSeekCoder-1.3B & 14.6 & 22.1 & 43.8 & 5.5 & 25.5 \\
Basic + No-code-prompt & DeepSeekCoder-1.3B & 15.7 & 21.3 & 37.5 & 4.8 & 24.4 \\
Basic + Reasoning-Step\&Code & DeepSeekCoder-1.3B & \textbf{17.8} & \textbf{25.5} & \textbf{59.4} & \textbf{5.9} & \textbf{26.1} \\
\bottomrule
\end{tabularx}

\label{tab:ablation_mathematical_code}
\end{table}

\textbf{Analysis of the impact of the mathematical code}. We analyze the impact of the mathematical code on continued pretraining by comparing the results of adding and not adding the mathematical code. As shown in Tab.~\ref{tab:ablation_mathematical_code}, the addition of the mathematical code in the pretraining corpus significantly improves performance across all five datasets. Note that the mathematical code only constitutes 14.1\% of the 19.2B tokens in the MathCode-Pile dataset, yet the improvement in accuracy it brings about compared to the total improvement in accuracy ($\frac{acc_\text{MathCode-Pile} - acc_\text{basic}}{acc_\text{MathCodePile} - acc_\text{orig}}$) is 21.8\%, 27.1\%, 44.5\%, 66.2\%, and 35.1\% on the five benchmark datasets, respectively, demonstrating the effectiveness of the mathematical code. Comparison across different training steps is shown in Appendix~\ref{sec:trend}.

We also analyze the design choice of concatenating the natural language reasoning step with the mathematical code for pretraining. This analysis is conducted by studying the results of adding only the natural language reasoning steps, and separately adding only the translated code. As shown in Tab.~\ref{tab:ablation_mathematical_code}, Basic + Reasoning-step-only represents adding only the natural language reasoning steps; Basic + Trans-code-only represents adding only the translated code; and Basic + Reasoning-Step\&Code represents adding the concatenated data that combines both. The Basic + Reasoning-Step\&Code configuration results in the best performance, demonstrating the importance of including both the natural language reasoning step and the translated mathematical code.

To rule out the possibility that the improvement comes from the higher quality of texts generated by Llama-3.1-70B-Instruct, we use a prompt that asks Llama-3.1-70B-Instruct to rewrite the given text. The details of this prompt are provided in Appendix~\ref{sec:simple_prompt}. We present the results of replacing the mathematical code with texts generated using this prompt in Tab.~\ref{tab:ablation_mathematical_code}, labeled as ``Basic + No-code-prompt''. Our method of generating mathematical code accompanied with corresponding reasoning steps outperforms this baseline, demonstrating the effectiveness of our approach.

\begin{table}[t]\fontsize{9}{11}\selectfont
\centering
\caption{Analysis of the effect of different components in MathCoder2-Pile. The base model is DeepSeekCoder-1.3B.}
\begin{tabularx}{\columnwidth}{>{\raggedright\arraybackslash\hsize=1.6\hsize}X  *{4}{>{\centering\arraybackslash\hsize=.28\hsize}X} >{\centering\arraybackslash\hsize=0.8\hsize}X}
\toprule
\textbf{Data Composition} & \textbf{MATH} &   \textbf{GSM8K}  & \textbf{SAT} &  \textbf{OCW} & \textbf{MMLU-MATH} \\
\midrule
No Math Training  & 4.8 & 4.3 & 18.8 & 2.6 & 24.8 \\
\midrule
filtered-OpenWebMath (4.8B)  & 9.0 & 11.4 & 34.4 & 3.7 & 25.4 \\
OpenWebMath (12.9B)  &  9.4 & 11.2 & 31.3 & 2.6 & 24.4 \\
\midrule
filtered-CC-En-math (6.4B)  & 9.1 & 12.1 & 31.3 & 3.7 & 25.2 \\
CC-En-math (22.1B)  & 8.4 & 13.0 & 25.0& 2.9 & 25.0 \\
\midrule
filtered-OpenWebMath + textbooks & 9.4 & 12.7 & 50.0 & 4.0 & 25.4 \\
filtered-OpenWebMath + synthetic data & 10.8 & 12.6 & 50.0 & 4.0 & 25.6 \\
filtered-OpenWebMath + code & 9.4 & 12.1 & 46.9 & 4.0 & 25.4 \\
\midrule
MathCoder2-Pile & \textbf{17.8} & \textbf{25.5} & \textbf{59.4} & \textbf{5.9} & \textbf{26.1} \\
\bottomrule
\end{tabularx}

\label{tab:ablation_basic}
\end{table}

\textbf{Analysis of the impact of various parts of the basic data}. We perform experiments on a smaller model, DeepSeekCoder-1.3B, using different parts of the basic data. As demonstrated in Tab.~\ref{tab:ablation_basic}, filtered-OpenWebMath and filtered-CC-En-math significantly improve the performance of the model. In comparison, textbooks, synthetic data, and code are smaller in data size and play a less important role. As each of these parts of data is too small for individual pretraining, we combine them with OpenWebMath-filtered to show that they each bring a small yet noticeable improvement compared to using only OpenWebMath-filtered. Since we performed filtering on OpenWebMath and the initially filtered CC-En to remove irrelevant data, we also compare the performance before and after filtering. We observe that there is no obvious degradation in performance after removing irrelevant content, showing the effectiveness of the filtering.

\begin{table}[t]\fontsize{9}{11}\selectfont
\centering
\caption{Comparison between MathCode-Pile and other Mathematical Pretrain datasets.}
\begin{tabularx}{\columnwidth}{>{\raggedright\arraybackslash\hsize=1.14\hsize}X >{\centering\arraybackslash\hsize=1.1\hsize}X *{4}{>{\centering\arraybackslash\hsize=.28\hsize}X} >{\centering\arraybackslash\hsize=0.8\hsize}X}
\toprule
\textbf{Pretrain Dataset} & \textbf{Base Model} & \textbf{MATH} &   \textbf{GSM8K}  & \textbf{SAT} &  \textbf{OCW} & \textbf{MMLU-MATH} \\
\midrule
No Math Training & DeepSeekCoder-1.3B & 4.8 & 4.3 & 18.8 & 2.6 & 24.8 \\
\midrule
OpenWebMath & DeepSeekCoder-1.3B & 9.4 & 11.2 & 31.3 & 2.6 & 24.4 \\
Proof-Pile-2 & DeepSeekCoder-1.3B & 9.2 & 11.2 & 50.0 & 4.4 & 25.8 \\
MathPile & DeepSeekCoder-1.3B & 5.3 & 3.4 & 21.9 & 2.2 & 24.9 \\
DeepSeekMath Corpus & DeepSeekLLM-1.3B & 13.6 & 23.8 & 56.3 & 4.8 & - \\
\midrule
MathCoder2-Pile & DeepSeekCoder-1.3B & \textbf{17.8} & \textbf{25.5} & \textbf{59.4} & \textbf{5.9} & \textbf{26.1} \\
\bottomrule
\end{tabularx}

\label{tab:ablation_other_corpora}
\end{table}

\textbf{Comparison with other open-source mathematical pretraining corpora}. We compare MathCode-Pile with various other open-source mathematical pretraining corpora using DeepSeekCoder-1.3B. We train each corpus for 3 epochs with a global batch size of 2 million tokens and a 4096 token context length, since we observe that the model's performance usually saturates around 3 epochs. As shown in Tab.~\ref{tab:ablation_other_corpora}, MathCode-Pile significantly outperforms OpenWebMath, Proof-Pile-2, and MathPile when trained on DeepSeekCoder-1.3B. The DeepSeekMath Corpus is not open-source, and its performance on DeepSeekLLM-1.3B is taken from~\cite{shao2024deepseekmathpushinglimitsmathematical}, which is trained for 150B tokens, more than our MathCode-Pile's training of approximately 60B tokens. The 1.3B model trained with MathCode-Pile outperforms the 1.3B model trained with DeepSeekMath Corpus.

\begin{table}[t]\fontsize{9}{11}\selectfont
\centering
\caption{Comparison between finetuning the original Llama-3-8B, MathCoder2-Basic-Llama-3-8B, and MathCoder2-Llama-3-8B on NuminaMath-TIR. MathCoder2-Basic-Llama-3-8B is the model resulting from continued pretraining on the basic data without adding the model-translated mathematical code.}
\begin{tabularx}{\columnwidth}{>{\raggedright\arraybackslash\hsize=1.3\hsize}X *{2}{>{\centering\arraybackslash\hsize=.4\hsize}X}  >{\centering\arraybackslash\hsize=.3\hsize}X >{\centering\arraybackslash\hsize=.6\hsize}X >{\centering\arraybackslash\hsize=0.5\hsize}X >{\centering\arraybackslash\hsize=.4\hsize}X}
\toprule
\textbf{Base Model} & \textbf{MATH} &   \textbf{GSM8K}  & \textbf{OCW} &  \textbf{Olympiad Bench}  & \textbf{SVAMP} \\
\midrule
Llama-3-8B & 56.1 & 80.1 & 24.6 & 28.4 &  83.8 \\
MathCoder2-Basic-Llama-3-8B & 62.9 & 81.3 & 26.8 & 32.9 &  86.7 \\
MathCoder2-Llama-3-8B & \textbf{65.1} & \textbf{84.5} & \textbf{34.6} & \textbf{34.4} &  \textbf{87.9} \\
\bottomrule
\end{tabularx}

\label{tab:ablation_post_training}
\end{table}

\textbf{Analysis of the improvement on the potential of being finetuned for TIR reasoning}. To analyze the effect of the model-translated mathematical code on LLMs' potential to be finetuned for TIR reasoning, we finetune the original Llama-3-8B, MathCoder2-Basic-Llama-3-8B, and MathCoder2-Llama-3-8B on NuminaMath-TIR\footnote{https://huggingface.co/datasets/AI-MO/NuminaMath-TIR} for three epochs, respectively. As shown in Tab.~\ref{tab:ablation_post_training}, the results of finetuning on MathCoder2-Basic-Llama-3-8B are higher than the results of finetuning on Llama-3-8B. Finetuning on MathCoder2-Llama-3-8B results in even higher performance than finetuning on MathCoder2-Basic-Llama-3-8B, showing that the addition of mathematical code effectively enhances the models' potential of being finetuned for TIR reasoning.

\section{Related Work}

\textbf{Continued mathematical pretraining.} Several works~\citep{shao2024deepseekmathpushinglimitsmathematical, azerbayev2024llemmaopenlanguagemodel,ying2024internlmmath, yang2024qwen25mathtechnicalreportmathematical} have explored the continued pretraining of LLMs on mathematical data, such as mathematical web content, synthetic data, and code. InternLM-Math~\citep{ying2024internlmmath} and Query of CC~\cite{fei2024queryccunearthinglarge} use BM25 for data retrieval, while other works such as DeepSeekMath~\citep{shao2024deepseekmathpushinglimitsmathematical} and Qwen2-Math~\citep{yang2024qwen25mathtechnicalreportmathematical} employ fastText~\citep{joulin2016bagtricksefficienttext} and other meta-information to retrieve texts from Common Crawl. Our approach follows these methods by using fastText for data filtering, and we introduce a second iteration of finer filtering to retain more relevant data. MathPile~\citep{wang2023generativeaimathi} and phi~\citep{gunasekar2023textbooksneed} utilize real or synthesized textbooks, while Llemma~\citep{azerbayev2024llemmaopenlanguagemodel} and Qwen2-Math~\citep{yang2024qwen25mathtechnicalreportmathematical} incorporate math-related code in their datasets. However, unlike our method of generating mathematical code with accompanied natural language reasoning, their code mostly has no natural language explanations or context. Our work builds on these prior efforts by collecting and expanding upon these sources of math-related text. Unlike works that only open-source their model weights, we take a more transparent approach by open-sourcing both our data processing and model training code, thereby ensuring reproducibility and facilitating future research in this field. Compared to Llemma~\citep{azerbayev2024llemmaopenlanguagemodel}, which also open-source their data and training code, our method results in better performance on mathematical reasoning tasks.

\textbf{Synthetic data.} Numerous finetuning~\citep{yu2024metamathbootstrapmathematicalquestions, wang2023mathcoderseamlesscodeintegration, lu2024mathgeniegeneratingsyntheticdata} and pretraining~\cite{gunasekar2023textbooksneed, wang2023generativeaimathi, yang2024qwen25mathtechnicalreportmathematical} studies have explored training on synthetic data generated using language models or predefined templates. MathGLM~\citep{yang2023gptsolvemathematicalproblems} and InternLM-Math~\citep{ying2024internlmmath} use templates to generate synthetic numerical operation data, while phi~\citep{gunasekar2023textbooksneed} produces textbook-quality data with models. EntiGraph~\citep{yang2024syntheticcontinuedpretraining} generates diverse text by drawing connections between sampled entities. Our work proposes a novel method for extracting mathematical reasoning steps and generating synthetic code snippets that captures the underlying reasoning processes.

\textbf{Post-training.} There are many methods for further improving the mathematical problem-solving abilities of LLMs. Supervised finetuning adjusts pretrained models using math problems and solutions in various formats, such as Chain-of-Thought~\citep{yu2024metamathbootstrapmathematicalquestions, yuan2023scalingrelationshiplearningmathematical}, Program-of-Thought~\citep{yue2023mammothbuildingmathgeneralist}, and Tool-Integrated Reasoning~\citep{gou2024toratoolintegratedreasoningagent, wang2023mathcoderseamlesscodeintegration, liao2024mariomathreasoningcode}. Reinforcement learning~\cite{lightman2023letsverifystepstep, wang2024mathshepherdverifyreinforcellms} and Direct Preference Optimization~\cite{rafailov2024directpreferenceoptimizationlanguage, xu2024chatglmmathimprovingmathproblemsolving, lu2024stepcontrolleddpoleveragingstepwise} utilize mathematical preference data to adjust the models' outputs. These methods are diverse and reveal the potential of pretrained models. Their performance is often influenced by the quality of the training data used in the pretraining stage. To explore the potential of finetuning our pretrained models for downstream tasks, we conduct supervised finetuning with existing open-source data.

\section{Limitations and Future Work}

One limitation of our work is that our continued pretraining corpus focuses primarily on mathematics and does not intentionally include other STEM subjects, such as physics and chemistry. Additionally, our pretraining data consists entirely of English texts, without incorporating math-related content in other languages, like Chinese. Due to limitations in computational resources, we only trained models ranging from 1.3B to 8B parameters. Future work could address these limitations by expanding the dataset to include other subjects and languages and by training on larger language models. Also, this paper primarily focuses on continued mathematical pretraining, so we did not apply reinforcement learning methods like PPO and GRPO, or Direct Preference Optimization in our post-training phase, which can further improve performance on mathematical reasoning tasks. In the future, we could explore these methods on our finetuned models.

\section{Conclusion}

In this paper, we present an effective open-source continued mathematical pretraining pipeline for enhancing mathematical reasoning of LLMs. Through the meticulous collection and filtering of diverse math-related texts, such as mathematical web content, synthetic data, code that uses mathematical packages, and math textbooks, we curate a basic dataset for continued mathematical pretraining. We then propose a novel method for extracting mathematical reasoning steps from the previously collected dataset and translating them to code snippets reflecting the underlying reasoning processes. By combining the basic data with the model-generated mathematical code accompanied with their corresponding reasoning steps, we produce a 19.2B-token mathematical pretraining corpus named MathCode-Pile, which significantly improves the performance of four different base models across five representative mathematical benchmarks. By open-sourcing the entire data processing pipeline and model training code, we actively promote transparency, reproducibility, and collaboration within the research community, facilitating future research in this area.

\bibliography{iclr2025_conference}

\begin{thebibliography}{38}
\providecommand{\natexlab}[1]{#1}
\providecommand{\url}[1]{\texttt{#1}}
\expandafter\ifx\csname urlstyle\endcsname\relax
  \providecommand{\doi}[1]{doi: #1}\else
  \providecommand{\doi}{doi: \begingroup \urlstyle{rm}\Url}\fi

\bibitem[Azerbayev et~al.(2024)Azerbayev, Schoelkopf, Paster, Santos, McAleer, Jiang, Deng, Biderman, and Welleck]{azerbayev2024llemmaopenlanguagemodel}
Zhangir Azerbayev, Hailey Schoelkopf, Keiran Paster, Marco~Dos Santos, Stephen McAleer, Albert~Q. Jiang, Jia Deng, Stella Biderman, and Sean Welleck.
\newblock Llemma: An open language model for mathematics, 2024.
\newblock URL \url{https://arxiv.org/abs/2310.10631}.

\bibitem[Cai et~al.(2024)Cai, Cao, Chen, Chen, Chen, Chen, Chen, Chen, Chen, Chu, Dong, Duan, Fan, Fei, Gao, Ge, Gu, Gu, Gui, Guo, Guo, He, Hu, Huang, Jiang, Jiao, Jin, Lei, Li, Li, Li, Li, Li, Li, Liu, Liu, Hong, Liu, Liu, Liu, Lv, Lv, Lv, Ma, Ma, Ma, Ning, Ouyang, Qiu, Qu, Shang, Shao, Song, Song, Sui, Sun, Sun, Tang, Wang, Wang, Wang, Wang, Wang, Wang, Wang, Wei, Weng, Wu, Xiong, Xu, Xu, Yan, Yan, Yang, Ye, Ying, Yu, Yu, Zang, Zhang, Zhang, Zhang, Zhang, Zhang, Zhang, Zhang, Zhang, Zhang, Zhang, Zhang, Zhao, Zhao, Zhao, Zhou, Zhou, Zhuo, Zou, Qiu, Qiao, and Lin]{cai2024internlm2}
Zheng Cai, Maosong Cao, Haojiong Chen, Kai Chen, Keyu Chen, Xin Chen, Xun Chen, Zehui Chen, Zhi Chen, Pei Chu, Xiaoyi Dong, Haodong Duan, Qi~Fan, Zhaoye Fei, Yang Gao, Jiaye Ge, Chenya Gu, Yuzhe Gu, Tao Gui, Aijia Guo, Qipeng Guo, Conghui He, Yingfan Hu, Ting Huang, Tao Jiang, Penglong Jiao, Zhenjiang Jin, Zhikai Lei, Jiaxing Li, Jingwen Li, Linyang Li, Shuaibin Li, Wei Li, Yining Li, Hongwei Liu, Jiangning Liu, Jiawei Hong, Kaiwen Liu, Kuikun Liu, Xiaoran Liu, Chengqi Lv, Haijun Lv, Kai Lv, Li~Ma, Runyuan Ma, Zerun Ma, Wenchang Ning, Linke Ouyang, Jiantao Qiu, Yuan Qu, Fukai Shang, Yunfan Shao, Demin Song, Zifan Song, Zhihao Sui, Peng Sun, Yu~Sun, Huanze Tang, Bin Wang, Guoteng Wang, Jiaqi Wang, Jiayu Wang, Rui Wang, Yudong Wang, Ziyi Wang, Xingjian Wei, Qizhen Weng, Fan Wu, Yingtong Xiong, Chao Xu, Ruiliang Xu, Hang Yan, Yirong Yan, Xiaogui Yang, Haochen Ye, Huaiyuan Ying, Jia Yu, Jing Yu, Yuhang Zang, Chuyu Zhang, Li~Zhang, Pan Zhang, Peng Zhang, Ruijie Zhang, Shuo Zhang, Songyang Zhang, Wenjian Zhang,
  Wenwei Zhang, Xingcheng Zhang, Xinyue Zhang, Hui Zhao, Qian Zhao, Xiaomeng Zhao, Fengzhe Zhou, Zaida Zhou, Jingming Zhuo, Yicheng Zou, Xipeng Qiu, Yu~Qiao, and Dahua Lin.
\newblock Internlm2 technical report, 2024.

\bibitem[Cobbe et~al.(2021)Cobbe, Kosaraju, Bavarian, Chen, Jun, Kaiser, Plappert, Tworek, Hilton, Nakano, Hesse, and Schulman]{cobbe2021trainingverifierssolvemath}
Karl Cobbe, Vineet Kosaraju, Mohammad Bavarian, Mark Chen, Heewoo Jun, Lukasz Kaiser, Matthias Plappert, Jerry Tworek, Jacob Hilton, Reiichiro Nakano, Christopher Hesse, and John Schulman.
\newblock Training verifiers to solve math word problems, 2021.
\newblock URL \url{https://arxiv.org/abs/2110.14168}.

\bibitem[Dubey et~al.(2024)Dubey, Jauhri, Pandey, Kadian, Al-Dahle, Letman, Mathur, Schelten, Yang, Fan, et~al.]{dubey2024llama}
Abhimanyu Dubey, Abhinav Jauhri, Abhinav Pandey, Abhishek Kadian, Ahmad Al-Dahle, Aiesha Letman, Akhil Mathur, Alan Schelten, Amy Yang, Angela Fan, et~al.
\newblock The llama 3 herd of models.
\newblock \emph{arXiv preprint arXiv:2407.21783}, 2024.

\bibitem[Fei et~al.(2024)Fei, Shao, Li, Zeng, He, Yan, Lin, and Qiu]{fei2024queryccunearthinglarge}
Zhaoye Fei, Yunfan Shao, Linyang Li, Zhiyuan Zeng, Conghui He, Hang Yan, Dahua Lin, and Xipeng Qiu.
\newblock Query of cc: Unearthing large scale domain-specific knowledge from public corpora, 2024.
\newblock URL \url{https://arxiv.org/abs/2401.14624}.

\bibitem[Gao et~al.(2024)Gao, Tow, Abbasi, Biderman, Black, DiPofi, Foster, Golding, Hsu, Le~Noac'h, Li, McDonell, Muennighoff, Ociepa, Phang, Reynolds, Schoelkopf, Skowron, Sutawika, Tang, Thite, Wang, Wang, and Zou]{eval-harness}
Leo Gao, Jonathan Tow, Baber Abbasi, Stella Biderman, Sid Black, Anthony DiPofi, Charles Foster, Laurence Golding, Jeffrey Hsu, Alain Le~Noac'h, Haonan Li, Kyle McDonell, Niklas Muennighoff, Chris Ociepa, Jason Phang, Laria Reynolds, Hailey Schoelkopf, Aviya Skowron, Lintang Sutawika, Eric Tang, Anish Thite, Ben Wang, Kevin Wang, and Andy Zou.
\newblock A framework for few-shot language model evaluation, 07 2024.
\newblock URL \url{https://zenodo.org/records/12608602}.

\bibitem[Gou et~al.(2024)Gou, Shao, Gong, Shen, Yang, Huang, Duan, and Chen]{gou2024toratoolintegratedreasoningagent}
Zhibin Gou, Zhihong Shao, Yeyun Gong, Yelong Shen, Yujiu Yang, Minlie Huang, Nan Duan, and Weizhu Chen.
\newblock Tora: A tool-integrated reasoning agent for mathematical problem solving, 2024.
\newblock URL \url{https://arxiv.org/abs/2309.17452}.

\bibitem[Gunasekar et~al.(2023)Gunasekar, Zhang, Aneja, Mendes, Giorno, Gopi, Javaheripi, Kauffmann, de~Rosa, Saarikivi, Salim, Shah, Behl, Wang, Bubeck, Eldan, Kalai, Lee, and Li]{gunasekar2023textbooksneed}
Suriya Gunasekar, Yi~Zhang, Jyoti Aneja, Caio César~Teodoro Mendes, Allie~Del Giorno, Sivakanth Gopi, Mojan Javaheripi, Piero Kauffmann, Gustavo de~Rosa, Olli Saarikivi, Adil Salim, Shital Shah, Harkirat~Singh Behl, Xin Wang, Sébastien Bubeck, Ronen Eldan, Adam~Tauman Kalai, Yin~Tat Lee, and Yuanzhi Li.
\newblock Textbooks are all you need, 2023.
\newblock URL \url{https://arxiv.org/abs/2306.11644}.

\bibitem[Hendrycks et~al.(2021{\natexlab{a}})Hendrycks, Burns, Basart, Zou, Mazeika, Song, and Steinhardt]{hendrycks2021measuringmassivemultitasklanguage}
Dan Hendrycks, Collin Burns, Steven Basart, Andy Zou, Mantas Mazeika, Dawn Song, and Jacob Steinhardt.
\newblock Measuring massive multitask language understanding, 2021{\natexlab{a}}.
\newblock URL \url{https://arxiv.org/abs/2009.03300}.

\bibitem[Hendrycks et~al.(2021{\natexlab{b}})Hendrycks, Burns, Kadavath, Arora, Basart, Tang, Song, and Steinhardt]{hendrycks2021measuringmathematicalproblemsolving}
Dan Hendrycks, Collin Burns, Saurav Kadavath, Akul Arora, Steven Basart, Eric Tang, Dawn Song, and Jacob Steinhardt.
\newblock Measuring mathematical problem solving with the math dataset, 2021{\natexlab{b}}.
\newblock URL \url{https://arxiv.org/abs/2103.03874}.

\bibitem[Jiang et~al.(2023)Jiang, Sablayrolles, Mensch, Bamford, Chaplot, de~las Casas, Bressand, Lengyel, Lample, Saulnier, Lavaud, Lachaux, Stock, Scao, Lavril, Wang, Lacroix, and Sayed]{jiang2023mistral7b}
Albert~Q. Jiang, Alexandre Sablayrolles, Arthur Mensch, Chris Bamford, Devendra~Singh Chaplot, Diego de~las Casas, Florian Bressand, Gianna Lengyel, Guillaume Lample, Lucile Saulnier, Lélio~Renard Lavaud, Marie-Anne Lachaux, Pierre Stock, Teven~Le Scao, Thibaut Lavril, Thomas Wang, Timothée Lacroix, and William~El Sayed.
\newblock Mistral 7b, 2023.
\newblock URL \url{https://arxiv.org/abs/2310.06825}.

\bibitem[Joulin et~al.(2016)Joulin, Grave, Bojanowski, and Mikolov]{joulin2016bagtricksefficienttext}
Armand Joulin, Edouard Grave, Piotr Bojanowski, and Tomas Mikolov.
\newblock Bag of tricks for efficient text classification, 2016.
\newblock URL \url{https://arxiv.org/abs/1607.01759}.

\bibitem[Lewkowycz et~al.(2022)Lewkowycz, Andreassen, Dohan, Dyer, Michalewski, Ramasesh, Slone, Anil, Schlag, Gutman-Solo, Wu, Neyshabur, Gur-Ari, and Misra]{lewkowycz2022solvingquantitativereasoningproblems}
Aitor Lewkowycz, Anders Andreassen, David Dohan, Ethan Dyer, Henryk Michalewski, Vinay Ramasesh, Ambrose Slone, Cem Anil, Imanol Schlag, Theo Gutman-Solo, Yuhuai Wu, Behnam Neyshabur, Guy Gur-Ari, and Vedant Misra.
\newblock Solving quantitative reasoning problems with language models, 2022.
\newblock URL \url{https://arxiv.org/abs/2206.14858}.

\bibitem[Li et~al.(2023)Li, Allal, Zi, Muennighoff, Kocetkov, Mou, Marone, Akiki, Li, Chim, Liu, Zheltonozhskii, Zhuo, Wang, Dehaene, Davaadorj, Lamy-Poirier, Monteiro, Shliazhko, Gontier, Meade, Zebaze, Yee, Umapathi, Zhu, Lipkin, Oblokulov, Wang, Murthy, Stillerman, Patel, Abulkhanov, Zocca, Dey, Zhang, Fahmy, Bhattacharyya, Yu, Singh, Luccioni, Villegas, Kunakov, Zhdanov, Romero, Lee, Timor, Ding, Schlesinger, Schoelkopf, Ebert, Dao, Mishra, Gu, Robinson, Anderson, Dolan-Gavitt, Contractor, Reddy, Fried, Bahdanau, Jernite, Ferrandis, Hughes, Wolf, Guha, von Werra, and de~Vries]{li2023starcodersourceyou}
Raymond Li, Loubna~Ben Allal, Yangtian Zi, Niklas Muennighoff, Denis Kocetkov, Chenghao Mou, Marc Marone, Christopher Akiki, Jia Li, Jenny Chim, Qian Liu, Evgenii Zheltonozhskii, Terry~Yue Zhuo, Thomas Wang, Olivier Dehaene, Mishig Davaadorj, Joel Lamy-Poirier, João Monteiro, Oleh Shliazhko, Nicolas Gontier, Nicholas Meade, Armel Zebaze, Ming-Ho Yee, Logesh~Kumar Umapathi, Jian Zhu, Benjamin Lipkin, Muhtasham Oblokulov, Zhiruo Wang, Rudra Murthy, Jason Stillerman, Siva~Sankalp Patel, Dmitry Abulkhanov, Marco Zocca, Manan Dey, Zhihan Zhang, Nour Fahmy, Urvashi Bhattacharyya, Wenhao Yu, Swayam Singh, Sasha Luccioni, Paulo Villegas, Maxim Kunakov, Fedor Zhdanov, Manuel Romero, Tony Lee, Nadav Timor, Jennifer Ding, Claire Schlesinger, Hailey Schoelkopf, Jan Ebert, Tri Dao, Mayank Mishra, Alex Gu, Jennifer Robinson, Carolyn~Jane Anderson, Brendan Dolan-Gavitt, Danish Contractor, Siva Reddy, Daniel Fried, Dzmitry Bahdanau, Yacine Jernite, Carlos~Muñoz Ferrandis, Sean Hughes, Thomas Wolf, Arjun Guha, Leandro von
  Werra, and Harm de~Vries.
\newblock Starcoder: may the source be with you!, 2023.
\newblock URL \url{https://arxiv.org/abs/2305.06161}.

\bibitem[Liao et~al.(2024)Liao, Luo, Li, Wu, and Fan]{liao2024mariomathreasoningcode}
Minpeng Liao, Wei Luo, Chengxi Li, Jing Wu, and Kai Fan.
\newblock Mario: Math reasoning with code interpreter output -- a reproducible pipeline, 2024.
\newblock URL \url{https://arxiv.org/abs/2401.08190}.

\bibitem[Lightman et~al.(2023)Lightman, Kosaraju, Burda, Edwards, Baker, Lee, Leike, Schulman, Sutskever, and Cobbe]{lightman2023letsverifystepstep}
Hunter Lightman, Vineet Kosaraju, Yura Burda, Harri Edwards, Bowen Baker, Teddy Lee, Jan Leike, John Schulman, Ilya Sutskever, and Karl Cobbe.
\newblock Let's verify step by step, 2023.
\newblock URL \url{https://arxiv.org/abs/2305.20050}.

\bibitem[Lu et~al.(2024{\natexlab{a}})Lu, Zhou, Ren, Wang, Shi, Pan, Zhan, and Li]{lu2024mathgeniegeneratingsyntheticdata}
Zimu Lu, Aojun Zhou, Houxing Ren, Ke~Wang, Weikang Shi, Junting Pan, Mingjie Zhan, and Hongsheng Li.
\newblock Mathgenie: Generating synthetic data with question back-translation for enhancing mathematical reasoning of llms, 2024{\natexlab{a}}.
\newblock URL \url{https://arxiv.org/abs/2402.16352}.

\bibitem[Lu et~al.(2024{\natexlab{b}})Lu, Zhou, Wang, Ren, Shi, Pan, Zhan, and Li]{lu2024stepcontrolleddpoleveragingstepwise}
Zimu Lu, Aojun Zhou, Ke~Wang, Houxing Ren, Weikang Shi, Junting Pan, Mingjie Zhan, and Hongsheng Li.
\newblock Step-controlled dpo: Leveraging stepwise error for enhanced mathematical reasoning, 2024{\natexlab{b}}.
\newblock URL \url{https://arxiv.org/abs/2407.00782}.

\bibitem[Paster et~al.(2023)Paster, Santos, Azerbayev, and Ba]{paster2023openwebmathopendatasethighquality}
Keiran Paster, Marco~Dos Santos, Zhangir Azerbayev, and Jimmy Ba.
\newblock Openwebmath: An open dataset of high-quality mathematical web text, 2023.
\newblock URL \url{https://arxiv.org/abs/2310.06786}.

\bibitem[Rafailov et~al.(2024)Rafailov, Sharma, Mitchell, Ermon, Manning, and Finn]{rafailov2024directpreferenceoptimizationlanguage}
Rafael Rafailov, Archit Sharma, Eric Mitchell, Stefano Ermon, Christopher~D. Manning, and Chelsea Finn.
\newblock Direct preference optimization: Your language model is secretly a reward model, 2024.
\newblock URL \url{https://arxiv.org/abs/2305.18290}.

\bibitem[Rozière et~al.(2024)Rozière, Gehring, Gloeckle, Sootla, Gat, Tan, Adi, Liu, Sauvestre, Remez, Rapin, Kozhevnikov, Evtimov, Bitton, Bhatt, Ferrer, Grattafiori, Xiong, Défossez, Copet, Azhar, Touvron, Martin, Usunier, Scialom, and Synnaeve]{roziere2024codellamaopenfoundation}
Baptiste Rozière, Jonas Gehring, Fabian Gloeckle, Sten Sootla, Itai Gat, Xiaoqing~Ellen Tan, Yossi Adi, Jingyu Liu, Romain Sauvestre, Tal Remez, Jérémy Rapin, Artyom Kozhevnikov, Ivan Evtimov, Joanna Bitton, Manish Bhatt, Cristian~Canton Ferrer, Aaron Grattafiori, Wenhan Xiong, Alexandre Défossez, Jade Copet, Faisal Azhar, Hugo Touvron, Louis Martin, Nicolas Usunier, Thomas Scialom, and Gabriel Synnaeve.
\newblock Code llama: Open foundation models for code, 2024.
\newblock URL \url{https://arxiv.org/abs/2308.12950}.

\bibitem[Shao et~al.(2024)Shao, Wang, Zhu, Xu, Song, Bi, Zhang, Zhang, Li, Wu, and Guo]{shao2024deepseekmathpushinglimitsmathematical}
Zhihong Shao, Peiyi Wang, Qihao Zhu, Runxin Xu, Junxiao Song, Xiao Bi, Haowei Zhang, Mingchuan Zhang, Y.~K. Li, Y.~Wu, and Daya Guo.
\newblock Deepseekmath: Pushing the limits of mathematical reasoning in open language models, 2024.
\newblock URL \url{https://arxiv.org/abs/2402.03300}.

\bibitem[Touvron et~al.(2023)Touvron, Martin, Stone, Albert, Almahairi, Babaei, Bashlykov, Batra, Bhargava, Bhosale, Bikel, Blecher, Ferrer, Chen, Cucurull, Esiobu, Fernandes, Fu, Fu, Fuller, Gao, Goswami, Goyal, Hartshorn, Hosseini, Hou, Inan, Kardas, Kerkez, Khabsa, Kloumann, Korenev, Koura, Lachaux, Lavril, Lee, Liskovich, Lu, Mao, Martinet, Mihaylov, Mishra, Molybog, Nie, Poulton, Reizenstein, Rungta, Saladi, Schelten, Silva, Smith, Subramanian, Tan, Tang, Taylor, Williams, Kuan, Xu, Yan, Zarov, Zhang, Fan, Kambadur, Narang, Rodriguez, Stojnic, Edunov, and Scialom]{touvron2023llama2openfoundation}
Hugo Touvron, Louis Martin, Kevin Stone, Peter Albert, Amjad Almahairi, Yasmine Babaei, Nikolay Bashlykov, Soumya Batra, Prajjwal Bhargava, Shruti Bhosale, Dan Bikel, Lukas Blecher, Cristian~Canton Ferrer, Moya Chen, Guillem Cucurull, David Esiobu, Jude Fernandes, Jeremy Fu, Wenyin Fu, Brian Fuller, Cynthia Gao, Vedanuj Goswami, Naman Goyal, Anthony Hartshorn, Saghar Hosseini, Rui Hou, Hakan Inan, Marcin Kardas, Viktor Kerkez, Madian Khabsa, Isabel Kloumann, Artem Korenev, Punit~Singh Koura, Marie-Anne Lachaux, Thibaut Lavril, Jenya Lee, Diana Liskovich, Yinghai Lu, Yuning Mao, Xavier Martinet, Todor Mihaylov, Pushkar Mishra, Igor Molybog, Yixin Nie, Andrew Poulton, Jeremy Reizenstein, Rashi Rungta, Kalyan Saladi, Alan Schelten, Ruan Silva, Eric~Michael Smith, Ranjan Subramanian, Xiaoqing~Ellen Tan, Binh Tang, Ross Taylor, Adina Williams, Jian~Xiang Kuan, Puxin Xu, Zheng Yan, Iliyan Zarov, Yuchen Zhang, Angela Fan, Melanie Kambadur, Sharan Narang, Aurelien Rodriguez, Robert Stojnic, Sergey Edunov, and Thomas
  Scialom.
\newblock Llama 2: Open foundation and fine-tuned chat models, 2023.
\newblock URL \url{https://arxiv.org/abs/2307.09288}.

\bibitem[Wang et~al.(2023{\natexlab{a}})Wang, Ren, Zhou, Lu, Luo, Shi, Zhang, Song, Zhan, and Li]{wang2023mathcoderseamlesscodeintegration}
Ke~Wang, Houxing Ren, Aojun Zhou, Zimu Lu, Sichun Luo, Weikang Shi, Renrui Zhang, Linqi Song, Mingjie Zhan, and Hongsheng Li.
\newblock Mathcoder: Seamless code integration in llms for enhanced mathematical reasoning, 2023{\natexlab{a}}.
\newblock URL \url{https://arxiv.org/abs/2310.03731}.

\bibitem[Wang et~al.(2024)Wang, Li, Shao, Xu, Dai, Li, Chen, Wu, and Sui]{wang2024mathshepherdverifyreinforcellms}
Peiyi Wang, Lei Li, Zhihong Shao, R.~X. Xu, Damai Dai, Yifei Li, Deli Chen, Y.~Wu, and Zhifang Sui.
\newblock Math-shepherd: Verify and reinforce llms step-by-step without human annotations, 2024.
\newblock URL \url{https://arxiv.org/abs/2312.08935}.

\bibitem[Wang et~al.(2023{\natexlab{b}})Wang, Xia, and Liu]{wang2023generativeaimathi}
Zengzhi Wang, Rui Xia, and Pengfei Liu.
\newblock Generative ai for math: Part i -- mathpile: A billion-token-scale pretraining corpus for math, 2023{\natexlab{b}}.
\newblock URL \url{https://arxiv.org/abs/2312.17120}.

\bibitem[Xu et~al.(2024)Xu, Liu, Liu, Hou, Li, Zhang, Wang, Zeng, Du, Zhao, Tang, and Dong]{xu2024chatglmmathimprovingmathproblemsolving}
Yifan Xu, Xiao Liu, Xinghan Liu, Zhenyu Hou, Yueyan Li, Xiaohan Zhang, Zihan Wang, Aohan Zeng, Zhengxiao Du, Wenyi Zhao, Jie Tang, and Yuxiao Dong.
\newblock Chatglm-math: Improving math problem-solving in large language models with a self-critique pipeline, 2024.
\newblock URL \url{https://arxiv.org/abs/2404.02893}.

\bibitem[Yang et~al.(2024{\natexlab{a}})Yang, Yang, Hui, Zheng, Yu, Zhou, Li, Li, Liu, Huang, et~al.]{yang2024qwen2}
An~Yang, Baosong Yang, Binyuan Hui, Bo~Zheng, Bowen Yu, Chang Zhou, Chengpeng Li, Chengyuan Li, Dayiheng Liu, Fei Huang, et~al.
\newblock Qwen2 technical report.
\newblock \emph{arXiv preprint arXiv:2407.10671}, 2024{\natexlab{a}}.

\bibitem[Yang et~al.(2024{\natexlab{b}})Yang, Zhang, Hui, Gao, Yu, Li, Liu, Tu, Zhou, Lin, Lu, Xue, Lin, Liu, Ren, and Zhang]{yang2024qwen25mathtechnicalreportmathematical}
An~Yang, Beichen Zhang, Binyuan Hui, Bofei Gao, Bowen Yu, Chengpeng Li, Dayiheng Liu, Jianhong Tu, Jingren Zhou, Junyang Lin, Keming Lu, Mingfeng Xue, Runji Lin, Tianyu Liu, Xingzhang Ren, and Zhenru Zhang.
\newblock Qwen2.5-math technical report: Toward mathematical expert model via self-improvement, 2024{\natexlab{b}}.
\newblock URL \url{https://arxiv.org/abs/2409.12122}.

\bibitem[Yang et~al.(2023)Yang, Ding, Lv, Jiang, He, Guo, Bai, and Tang]{yang2023gptsolvemathematicalproblems}
Zhen Yang, Ming Ding, Qingsong Lv, Zhihuan Jiang, Zehai He, Yuyi Guo, Jinfeng Bai, and Jie Tang.
\newblock Gpt can solve mathematical problems without a calculator, 2023.
\newblock URL \url{https://arxiv.org/abs/2309.03241}.

\bibitem[Yang et~al.(2024{\natexlab{c}})Yang, Band, Li, Candès, and Hashimoto]{yang2024syntheticcontinuedpretraining}
Zitong Yang, Neil Band, Shuangping Li, Emmanuel Candès, and Tatsunori Hashimoto.
\newblock Synthetic continued pretraining, 2024{\natexlab{c}}.
\newblock URL \url{https://arxiv.org/abs/2409.07431}.

\bibitem[Ying et~al.(2024)Ying, Zhang, Li, Zhou, Shao, Fei, Ma, Hong, Liu, Wang, Wang, Wu, Li, Zhou, Liu, Zhang, Zhang, Yan, Qiu, Wang, Chen, and Lin]{ying2024internlmmath}
Huaiyuan Ying, Shuo Zhang, Linyang Li, Zhejian Zhou, Yunfan Shao, Zhaoye Fei, Yichuan Ma, Jiawei Hong, Kuikun Liu, Ziyi Wang, Yudong Wang, Zijian Wu, Shuaibin Li, Fengzhe Zhou, Hongwei Liu, Songyang Zhang, Wenwei Zhang, Hang Yan, Xipeng Qiu, Jiayu Wang, Kai Chen, and Dahua Lin.
\newblock Internlm-math: Open math large language models toward verifiable reasoning, 2024.

\bibitem[Yu et~al.(2024)Yu, Jiang, Shi, Yu, Liu, Zhang, Kwok, Li, Weller, and Liu]{yu2024metamathbootstrapmathematicalquestions}
Longhui Yu, Weisen Jiang, Han Shi, Jincheng Yu, Zhengying Liu, Yu~Zhang, James~T. Kwok, Zhenguo Li, Adrian Weller, and Weiyang Liu.
\newblock Metamath: Bootstrap your own mathematical questions for large language models, 2024.
\newblock URL \url{https://arxiv.org/abs/2309.12284}.

\bibitem[Yuan et~al.(2023)Yuan, Yuan, Li, Dong, Lu, Tan, Zhou, and Zhou]{yuan2023scalingrelationshiplearningmathematical}
Zheng Yuan, Hongyi Yuan, Chengpeng Li, Guanting Dong, Keming Lu, Chuanqi Tan, Chang Zhou, and Jingren Zhou.
\newblock Scaling relationship on learning mathematical reasoning with large language models, 2023.
\newblock URL \url{https://arxiv.org/abs/2308.01825}.

\bibitem[Yue et~al.(2023)Yue, Qu, Zhang, Fu, Huang, Sun, Su, and Chen]{yue2023mammothbuildingmathgeneralist}
Xiang Yue, Xingwei Qu, Ge~Zhang, Yao Fu, Wenhao Huang, Huan Sun, Yu~Su, and Wenhu Chen.
\newblock Mammoth: Building math generalist models through hybrid instruction tuning, 2023.
\newblock URL \url{https://arxiv.org/abs/2309.05653}.

\bibitem[Yue et~al.(2024)Yue, Zheng, Zhang, and Chen]{yue2024mammoth2scalinginstructionsweb}
Xiang Yue, Tuney Zheng, Ge~Zhang, and Wenhu Chen.
\newblock Mammoth2: Scaling instructions from the web, 2024.
\newblock URL \url{https://arxiv.org/abs/2405.03548}.

\bibitem[Zhang et~al.(2024)Zhang, Qu, Liu, Zhang, Lin, Yu, Pan, Cheng, Liu, Lin, Yuan, Zheng, Pang, Du, Liang, Ma, Li, Ma, Lin, Benetos, Yang, Zhou, Ma, Liu, Niu, Wang, Que, Liu, Liu, Guo, Gao, Zhou, Zhang, Zhou, Wang, Bai, Zhang, Zhang, Wang, Yang, Zhao, Zhang, Ouyang, Huang, and Chen]{zhang2024mapneohighlycapabletransparent}
Ge~Zhang, Scott Qu, Jiaheng Liu, Chenchen Zhang, Chenghua Lin, Chou~Leuang Yu, Danny Pan, Esther Cheng, Jie Liu, Qunshu Lin, Raven Yuan, Tuney Zheng, Wei Pang, Xinrun Du, Yiming Liang, Yinghao Ma, Yizhi Li, Ziyang Ma, Bill Lin, Emmanouil Benetos, Huan Yang, Junting Zhou, Kaijing Ma, Minghao Liu, Morry Niu, Noah Wang, Quehry Que, Ruibo Liu, Sine Liu, Shawn Guo, Soren Gao, Wangchunshu Zhou, Xinyue Zhang, Yizhi Zhou, Yubo Wang, Yuelin Bai, Yuhan Zhang, Yuxiang Zhang, Zenith Wang, Zhenzhu Yang, Zijian Zhao, Jiajun Zhang, Wanli Ouyang, Wenhao Huang, and Wenhu Chen.
\newblock Map-neo: Highly capable and transparent bilingual large language model series, 2024.
\newblock URL \url{https://arxiv.org/abs/2405.19327}.

\bibitem[Zhou et~al.(2024)Zhou, Wang, Lu, Shi, Luo, Qin, Lu, Jia, Song, Zhan, and Li]{zhou2024solving}
Aojun Zhou, Ke~Wang, Zimu Lu, Weikang Shi, Sichun Luo, Zipeng Qin, Shaoqing Lu, Anya Jia, Linqi Song, Mingjie Zhan, and Hongsheng Li.
\newblock Solving challenging math word problems using {GPT}-4 code interpreter with code-based self-verification.
\newblock In \emph{The Twelfth International Conference on Learning Representations}, 2024.
\newblock URL \url{https://openreview.net/forum?id=c8McWs4Av0}.

\end{thebibliography}
\bibliographystyle{iclr2025_conference}

\appendix

\section{Prompt for Annotation of Math Web Documents}
\label{sec:classify_prompt}

\begin{figure}[t!]
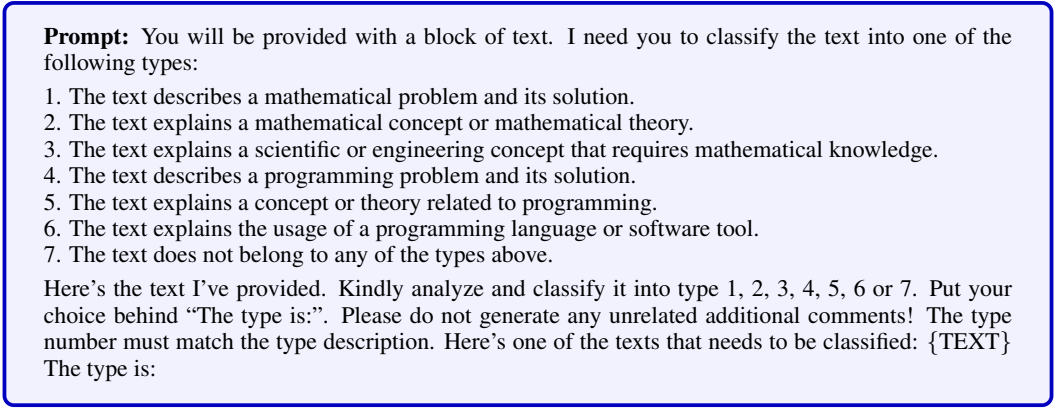

\begin{tcolorbox}[colback=blue!5!white,colframe=blue!75!black]
\begin{small}
\textbf{Prompt:}
You will be provided with a block of text. I need you to classify the text into one of the following types:
\vspace{1mm}

1. The text describes a mathematical problem and its solution.

2. The text explains a mathematical concept or mathematical theory.

3. The text explains a scientific or engineering concept that requires mathematical knowledge.

4. The text describes a programming problem and its solution.

5. The text explains a concept or theory related to programming.

6. The text explains the usage of a programming language or software tool.

7. The text does not belong to any of the types above.
\vspace{1mm}

Here's the text I've provided. Kindly analyze and classify it into type 1, 2, 3, 4, 5, 6 or 7. Put your choice behind ``The type is:''. Please do not generate any unrelated additional comments! The type number must match the type description. Here's one of the texts that needs to be classified: \{TEXT\} The type is:

\end{small}
\end{tcolorbox}
\caption{The prompt for annotation of OpenWebMath and the initially filtered CC-En documents. \{TEXT\} is replaced with the content of the document.}
\label{fig:prompt_filter}
\end{figure}

In this section, we present the prompt we used for annotation of documents in OpenWebMath and the initially filtered CC-En. The prompt, as shown in Fig.~\ref{fig:prompt_filter}, asks the model to classify the document into one of seven types, which are types of documents that frequently appear in the datasets. We observe that this method helps the model to better identify and filter out irrelevant text than using a binary classification of whether the text is related to math.

\section{Training Details of FastText Classifiers}
\label{sec:fasttext}

We employ an open-source library\footnote{https://fasttext.cc/} for training, configuring the vector dimension to 50, the learning rate to 0.5, the maximum length of word n-grams to 2, and the number of training epochs to 5. For the initial filtering of the Common Crawl corpus, we sample 3 million data points from the seed corpus of filtered-OpenWebMath as positive training examples and another 8 million web pages from Common Crawl as negative examples. For finer filtering, we use 2 million data points annotated as math-related by Mixtral-8x7B-Instruct as positive training samples and 1 million data points annotated as unrelated to math as negative training samples.

\section{Examples of the Model-generated Mathematical Code}
\label{sec:examples_translated_code}

This session shows several examples of the translation from math-related texts to mathematical code acompanied with corresponding reasoning steps. As shown in Tab.~\ref{tab:example_translation_1}, Tab.~\ref{tab:example_translation_2}, Tab.~\ref{tab:example_translation_3} and  Tab.~\ref{tab:example_translation_4}, the model first extract the LaTex expression alone with its conditions and result from the original text, then generates an Python code snippet based on this information.

\begin{table*}[t!]
    \centering
    \begin{tabular}{p{0.95\textwidth}}
    \toprule
\textbf{Original Text}:\\
Bayes and his Theorem\\
My earlier post on Bayesian probability seems to have generated quite a lot of readers, so this lunchtime I thought I’d add a little bit of background.\\
......\\
If we ask the question “what is the probability of exactly x successes from the possible n?”, the answer is given by the binomial distribution:\\
$$P_n(x|n,p)= C(n,x) p^x (1-p)^{n-x}$$
......\\
    \midrule
\textbf{Translation Output}:\\
Conditions Needed:\\
1. The number of trials (n) is a positive integer.\\
2. The probability of success (p) is a real number between 0 and 1.\\
3. The number of successes (x) is a non-negative integer less than or equal to n.\\

Computation Expression:\\
$P_n(x|n,p)= C(n,x) p^x (1-p)^{n-x}$

Computation Result:\\
The probability of exactly x successes in n independent trials, each with a probability of success p.\\

Python Code Snippet:
     \lstset{style=mystyle}
     \begin{lstlisting}[language=Python]
```python
import math

def binomial_distribution(n, x, p):
    combinations = math.comb(n, x)
    probability = combinations * (p ** x) * ((1 - p) ** (n - x))
    return probability

# Example usage:
n = 10  # number of trials
x = 5   # number of successes
p = 0.5  # probability of success
result = binomial_distribution(n, x, p)
print(result)
```
     \end{lstlisting} 
\\
     \bottomrule
    \end{tabular}
    \caption{An example of translating a piece of text to mathematical code and a corresponding reasoning step.}
    \label{tab:example_translation_1}
\end{table*}

\begin{table*}[t!]
    \centering
    \begin{tabular}{p{0.95\textwidth}}
    \toprule
\textbf{Original Text}:\\
\# Probability of getting 2 Aces, 2 Kings and 1 Queen in a five card poker hand (Part II)\\
So I reworked my formula in method 1 after getting help with my original question ......\\
$$\frac{{4 \choose 2} {4 \choose 2}{4 \choose 1}}{{52 \choose 5}} = \frac{3}{54145}$$
$$5.540678 * 10^{-5}$$
......\\
    \midrule
\textbf{Translation Output}:\\
Conditions Needed:\\
1. The total number of cards in a deck is 52.\\
2. We are drawing 5 cards from the deck.\\
3. We want to calculate the probability of getting 2 Aces, 2 Kings, and 1 Queen.\\

Computation Expression:\\
$$\frac{{4 \choose 2} {4 \choose 2}{4 \choose 1}}{{52 \choose 5}}$$

Computation Result:\\
$$5.540678 * 10^{-5}$$

Python Code Snippet:
     \lstset{style=mystyle}
     \begin{lstlisting}[language=Python]
```python
import math

def combination(n, k):
    return math.comb(n, k)

numerator = combination(4, 2) * combination(4, 2) * combination(4, 1)
denominator = combination(52, 5)

result = numerator / denominator
print(result)
```
     \end{lstlisting} 
\\
     \bottomrule
    \end{tabular}
    \caption{An example of translating a piece of text to mathematical code and a corresponding reasoning step.}
    \label{tab:example_translation_2}
\end{table*}

\begin{table*}[t!]
    \centering
    \begin{tabular}{p{0.95\textwidth}}
    \toprule
\textbf{Original Text}:\\
\# How do you use the product rule to differentiate $g(x)=(x^2+1)(x^2-2x)$?\\
......\\
$\text{and } h \left(x\right) = {x}^{2} - 2 x \Rightarrow h ' \left(x\right) = 2 x - 2$\\
......\\
    \midrule
\textbf{Translation Output}:\\
Conditions Needed:\\
1. $h(x)$ is a polynomial function\\
2. The derivative of $x^2$ is $2x$\\
3. The derivative of $-2x$ is $-2$\\
Computation Expression:\\
$\frac{d}{dx} (x^2 - 2x) = 2x - 2$\\
Computation Result:\\
$h'(x) = 2x - 2$\\
Python Code Snippet:
     \lstset{style=mystyle}
     \begin{lstlisting}[language=Python]
```python
import sympy as sp

x = sp.symbols('x')
h = x**2 - 2*x
h_prime = sp.diff(h, x)
print(h_prime)  # Output: 2*x - 2
```
     \end{lstlisting} 
\\
     \bottomrule
    \end{tabular}
    \caption{An example of translating a piece of text to mathematical code and a corresponding reasoning step.}
    \label{tab:example_translation_3}
\end{table*}

\begin{table*}[t!]
    \centering
    \begin{tabular}{p{0.95\textwidth}}
    \toprule
\textbf{Original Text}:\\
\# How do you use the product rule to differentiate $g(x)=(x^2+1)(x^2-2x)$?\\
......\\
$\text{and } h \left(x\right) = {x}^{2} - 2 x \Rightarrow h ' \left(x\right) = 2 x - 2$\\
......\\
    \midrule
\textbf{Translation Output}:\\
Conditions Needed:\\
1. $h(x)$ is a polynomial function\\
2. The derivative of $x^2$ is $2x$\\
3. The derivative of $-2x$ is $-2$\\
Computation Expression:\\
$\frac{d}{dx} (x^2 - 2x) = 2x - 2$\\
Computation Result:\\
$h'(x) = 2x - 2$\\
Python Code Snippet:
     \lstset{style=mystyle}
     \begin{lstlisting}[language=Python]
```python
import sympy as sp

x = sp.symbols('x')
h = x**2 - 2*x
h_prime = sp.diff(h, x)
print(h_prime)  # Output: 2*x - 2
```
     \end{lstlisting} 
\\
     \bottomrule
    \end{tabular}
    \caption{An example of translating a piece of text to mathematical code and a corresponding reasoning step.}
    \label{tab:example_translation_4}
\end{table*}

\section{Examples of Removed Irrelevant Texts}
\label{sec:irrelevant_texts}

In this section, we present several examples in the original OpenWebMath dataset that are irrelevant to mathematical reasoning and removed in the filtering process. As shown in Tab.~\ref{tab:example_irrelevant_1}, Tab.~\ref{tab:example_irrelevant_2}, and Tab.~\ref{tab:example_irrelevant_3}, the content of these documents are not related to math, but instead are about subjects such as politics, testing software, or web development. Removing these irrelevant texts have no obvious impact on the mathematical continued pretraining performance. 

\begin{table*}[t!]
    \centering
    \begin{tabular}{p{0.95\textwidth}}
    \toprule
\#\# Avoiding Weimar Russia

Matthew Yglesias writes:

Matthew Yglesias: Beyond Economics: Over at Brad DeLong's site you can see a fascinating discussion of America's Russia policy in the 1990s between DeLong, Martin Wolf, and Lawrence Summers. One remark I would make is that to an extraordinary extent, all three participants are willing to accept the premise that the only goal of US policy toward Russia in the 1990s was a good-faith effort to induce Russian prosperity, with such efforts being hampered by political constraints, the objective difficulty of the task, and pure policy errors...

Well, yes. Russia was once a superpower and may be one again. One would have thought that the history of 1914-1945 would teach ample lessons about the national security undesirability of trying to keep great powers--like Weimar Germany--poor and weak. One would have thought that the history of 1945-1990 would teach ample lessons about the national security desirability of trying to help great powers--like Japan and West Germany--become prosperous, democratic, and well-integrated into the world economy.

One top of the national-security strategic argument there is the economic argument: the fact that richer trading partners are better trading partners: they make more and more interesting stuff for us to buy.\\
......\\
     \bottomrule
    \end{tabular}
    \caption{An example of removed text irrelevant to mathematical reasoning in OpenWebMath.}
    \label{tab:example_irrelevant_1}
\end{table*}

\begin{table*}[t!]
    \centering
    \begin{tabular}{p{0.95\textwidth}}
    \toprule
\# MicroEJ Test Suite Engine¶

\#\# Introduction¶

The MicroEJ Test Suite Engine is a generic tool made for validating any development project using automatic testing.

This section details advanced configuration for users who wish to integrate custom test suites in their build flow.

The MicroEJ Test Suite Engine allows the user to test any kind of projects within the configuration of a generic Ant file.

The MicroEJ Test Suite Engine is already pre-configured for running test suites on a MicroEJ Platform (either on Simulator or on Device).

\#\# Using the MicroEJ Test Suite Ant Tasks¶

Multiple Ant tasks are available in the testsuite-engine.jar provided in the Build Kit:

• testsuite allows the user to run a given test suite and to retrieve an XML report document in a JUnit format.

• javaTestsuite is a subtask of the testsuite task, used to run a specialized test suite for Java (will only run Java classes).

• htmlReport is a task which will generate an HTML report from a list of JUnit report files.\\
......\\
     \bottomrule
    \end{tabular}
    \caption{An example of removed text irrelevant to mathematical reasoning in OpenWebMath.}
    \label{tab:example_irrelevant_2}
\end{table*}

\begin{table*}[t!]
    \centering
    \begin{tabular}{p{0.95\textwidth}}
    \toprule
By Kimserey Lam with

\# Conemu A Better Command Prompt For Windows

Jul 22nd, 2017 - written by Kimserey with .

When developing multiple Web api under multiple Visual Studio solutions, it can become very tedious to maintain, run and debug. Opening multiple instances of Visual Studio is very costly in term of memory and running all at once also clutter the screen which rapidly becomes irritating. With the advent of dotnet CLI tools, it has been clear that the next step would be to move out of the common “right click/build, F5” of Visual Studio and toward “dotnet run” on a command prompt. Last month I was looking for a Windows alternative of the bash terminal which can be found on Mac and I found ConEmu. ConEmu provides access to all typical shells via an enhanced UI. Today we will see how we can use ConEmu to ease our development process by leveraging only 2 of its features; the tasks and environment setup.

1. dotnet CLI
2. Setup environment
4. Apply to multiple services\\
......\\
     \bottomrule
    \end{tabular}
    \caption{An example of removed text irrelevant to mathematical reasoning in OpenWebMath.}
    \label{tab:example_irrelevant_3}
\end{table*}

\section{Prompt for Simple Rewriting to Improve Quality for Ablation Study}
\label{sec:simple_prompt}

To rule out the possibility that the improvement results solely from the enhanced quality of the texts generated by Llama-3.1-70B-Instruct, we designed a prompt asking Llama-3.1-70B-Instruct to rewrite the text, checking for mistakes in content and format to enhance accuracy and clarity, as shown in Fig.~\ref{fig:prompt_simple}.

\begin{figure}[t!]
\begin{tcolorbox}[colback=blue!5!white,colframe=blue!75!black]
\begin{small}
\textbf{Prompt:}
You will be presented with a text related to math. I need you to carefully read through the text. If you find any incorrect statments, erroneous computation steps, spelling mistakes, grammatical errors, or formatting issues, adjust them so that the error is corrected. Rewrite the text to make it more accurate and easier to understand. You should only output an adjusted version of the given text. Also, do not change the original language. Please do not generate any unrelated additional comments! The text is as follows: \{TEXT\}

You should output:
\end{small}
\end{tcolorbox}
\caption{The prompt asking Llama-3.1-70B-Instruct to simply rewrite the text and improve its quality. \{TEXT\} is replaced with the content of the document.}
\label{fig:prompt_simple}
\end{figure}

\section{Comparison Between Adding and Not Adding Mathematical Code}
\label{sec:trend}

\begin{figure*}[t]
    \centering
    \includegraphics[width=1.0\textwidth]{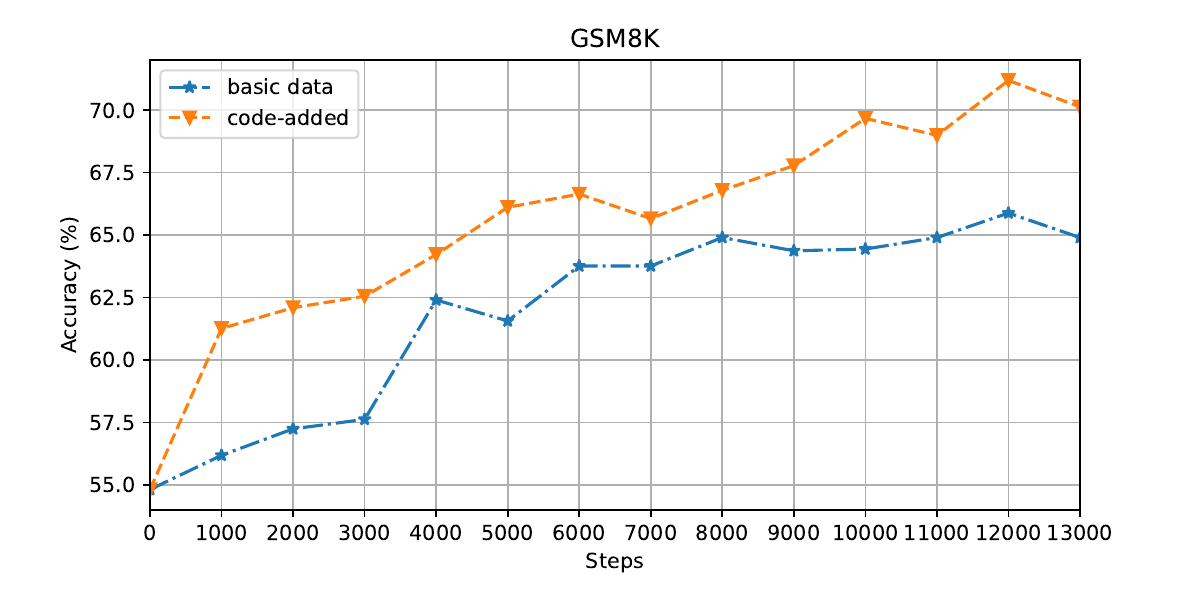}
    \caption{Comparison of the accuracy on GSM8K between adding and not adding mathematical code across different training steps.}
    
\label{fig:trend_gsm8k}
\end{figure*}

\begin{figure*}[t]
    \centering
    \includegraphics[width=1.0\textwidth]{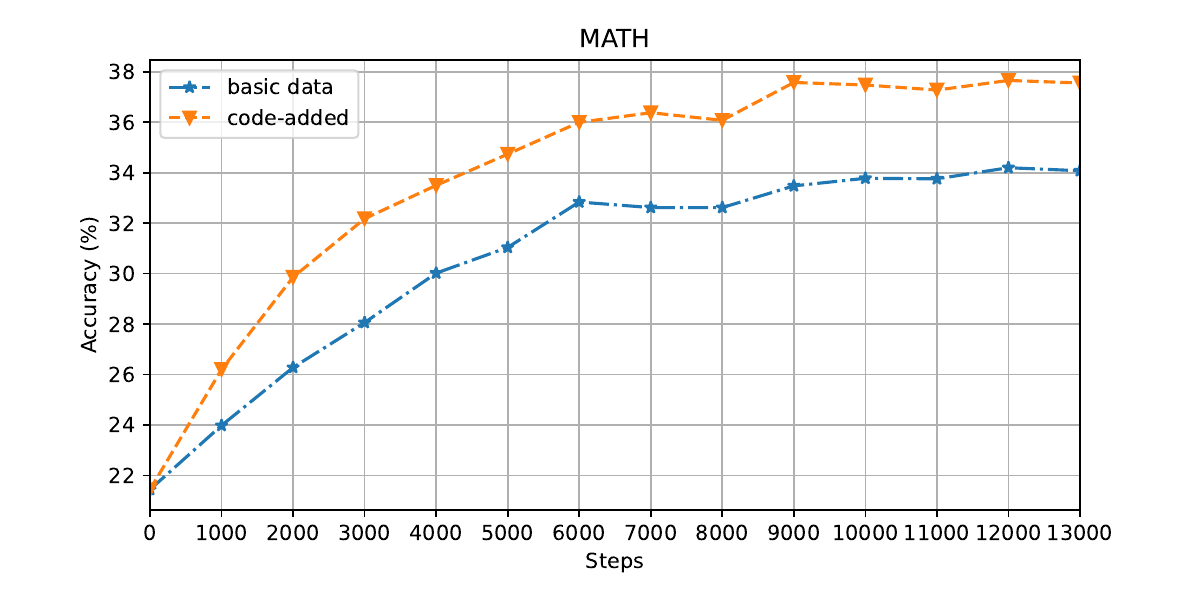}
    \caption{Comparison of the accuracy on MATH between adding and not adding mathematical code across different training steps.}
    
\label{fig:trend_math}
\end{figure*}

In this section, we present the comparison between adding and not adding mathematical code across different training steps. The experiments are conducted on Llama-3 8B. As shown in Fig.~\ref{fig:trend_gsm8k} and Fig.~\ref{fig:trend_math}, adding the model-translated mathematical code improves accuracy across different training steps.

\end{document}